\documentclass[draftclsnofoot,onecolumn,11pt]{IEEEtran}

\usepackage{amsmath}
\usepackage{amsfonts}
\usepackage{amssymb}
\usepackage{amsthm}
\usepackage{color}
\usepackage{array}
\usepackage{cite}
\usepackage{enumerate}
\usepackage{graphicx}
\usepackage[hang]{subfigure}
\usepackage{caption}
\usepackage{epstopdf}
\usepackage{epsfig}
\usepackage{footmisc}
\usepackage{amsbsy}
\usepackage{bm}
\usepackage{cases}
\usepackage{multirow}
\usepackage{algorithmic}
\usepackage{comment}
\usepackage{times}
\usepackage{paralist}
\usepackage[ruled]{algorithm2e}
\usepackage{setspace}
\usepackage[breakable]{tcolorbox}

\newcommand{\bmx}{{\bm x}}
\newcommand{\bmh}{{\bm h}}
\newcommand{\bms}{{\bm s}}

\newcommand{\bme}{{\bm e}}

\newcommand{\bmtheta}{{\bm \theta}}

\newcommand{\bmphi}{{\bm \phi}}

\newcommand{\overrs}{\overrightarrow{s}}

\newcommand{\overls}{\overleftarrow{s}}

\newcommand{\overrx}{\overrightarrow{x}}

\newcommand{\overlx}{\overleftarrow{x}}

\newcommand{\textX}{\text{X}}
\newcommand{\textH}{\text{H}}

\newcommand{\set}[1]{\ensuremath{\mathcal #1}}
\newcommand{\grad}[1]{\nabla #1}
\newcommand{\note}[1]{[\textcolor{red}{\textit{#1}}]}

\newcounter{aalign}
  \newcounter{thephase} 
\newcounter{thesubphase}[thephase] 

\newcounter{rtaskno}

\allowdisplaybreaks

\begin{document}
\title{An Introduction to \\  Probabilistic Spiking Neural Networks}
\author{
\IEEEauthorblockN{Hyeryung Jang, Osvaldo Simeone, Brian Gardner, and Andr\'e Gr\"uning}
}


\maketitle
\vspace{-1.5cm}
\begin{abstract}
Spiking neural networks (SNNs) are distributed trainable systems whose computing elements, or neurons, are characterized by internal analog dynamics and by digital and sparse synaptic communications. The sparsity of the synaptic spiking inputs and the corresponding event-driven nature of neural processing can be leveraged by energy-efficient hardware implementations, which can offer significant energy reductions as compared to conventional artificial neural networks (ANNs). The design of training algorithms lags behind the hardware implementations. Most existing training algorithms for SNNs have been designed either for biological plausibility or through conversion from pretrained ANNs via rate encoding. This article provides an introduction to SNNs by focusing on a probabilistic signal processing methodology that enables the direct derivation of learning rules by leveraging the unique time-encoding capabilities of SNNs. We adopt discrete-time probabilistic models for networked spiking neurons and derive supervised and unsupervised learning rules from first principles via variational inference. Examples and open research problems are also provided. \end{abstract}



\IEEEpeerreviewmaketitle

\section*{Introduction}

ANNs have become the de facto standard tool to carry out supervised, unsupervised, and reinforcement learning tasks. Their recent successes range from image classifiers that outperform human experts in medical diagnosis to machines that defeat professional players at complex games such as Go. These breakthroughs have built upon various algorithmic advances but have also heavily relied on the unprecedented availability of computing power and memory in data centers and cloud computing platforms. The resulting considerable energy requirements run counter to the constraints imposed by implementations on low-power mobile or embedded devices for such applications as personal health monitoring or neural prosthetics \cite{welling18:icmltalk}. 

\textbf{ANNs versus SNNs.} Various new hardware solutions have recently emerged that attempt to improve the energy efficiency of ANNs as inference machines by trading complexity for accuracy in the implementation of matrix operations. A different line of research, which is the subject of this article, seeks an alternative framework that enables efficient {\em online inference and learning} by taking inspiration from the working of the human brain. 

\begin{figure}[t!]
\centering
\subfigure[]{
\includegraphics[height=0.2\columnwidth]{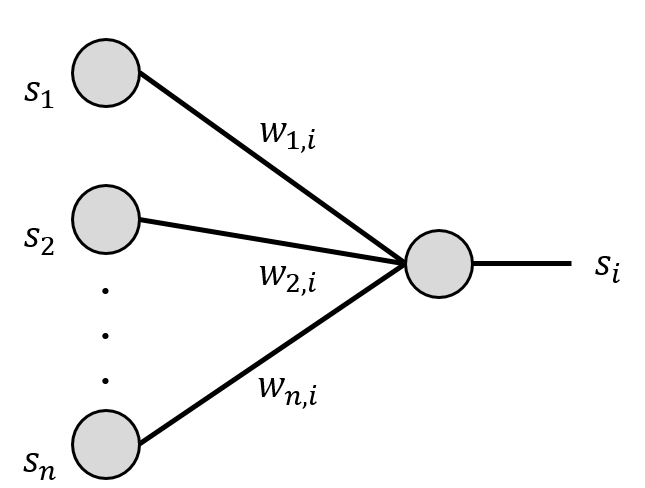}
}
\subfigure[]{
\includegraphics[height=0.2\columnwidth]{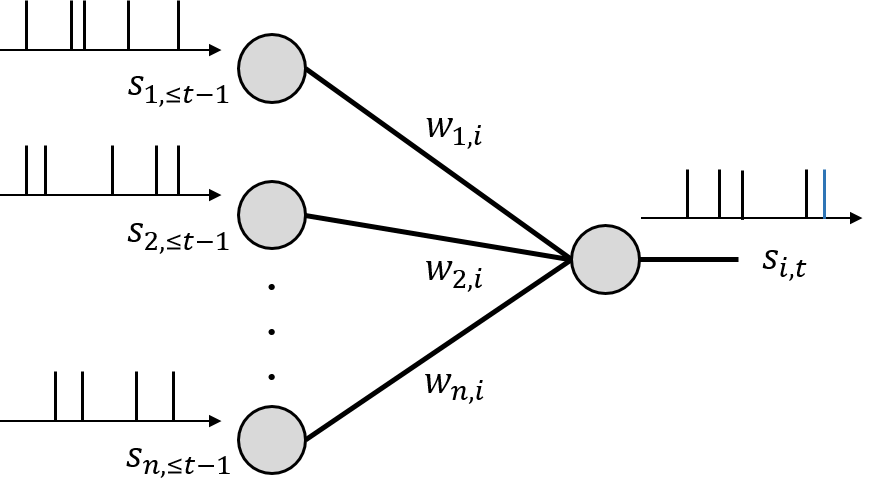}
}
  \caption{Illustration of NNs: (a) an ANN, where each neuron $i$ processes real numbers $s_1,\ldots,s_n$ to output and communicates a real number $s_i$ as a static nonlinearity and (b) an SNN, where dynamic spiking neurons process and communicate sparse spiking signals over time $t$ in a causal manner to output and communicate a binary spiking signal $s_{i,t}$. }
  \label{fig:ann_vs_snn}
  \vspace{-0.55cm}
\end{figure}

The human brain is capable of performing general and complex tasks via continuous adaptation at a minute fraction of the power required by state-of-the-art supercomputers and ANN-based models \cite{paugam2012computing}. Neurons in the human brain are qualitatively different from those in an ANN. They are \emph{dynamic} devices featuring recurrent behavior, rather than static nonlinearities, and they process and communicate using \emph{sparse spiking signals} over time, rather than real numbers. Inspired by this observation, as illustrated in Fig.~\ref{fig:ann_vs_snn}, SNNs have been introduced in the theoretical neuroscience literature as networks of dynamic spiking neurons \cite{maass1997networks}. SNNs have the unique capability to process information encoded in the timing of events, or spikes. Spikes are also used for synaptic communications, with synapses delaying and filtering signals before they reach the postsynaptic neuron. Because of the presence of synaptic delays, neurons in an SNN can be naturally connected via arbitrary recurrent topologies, unlike standard multilayer ANNs or chain-like recurrent neural networks.

Proof-of-concept and commercial hardware implementations of SNNs have demonstrated orders-of-magnitude improvements in terms of energy efficiency over ANNs \cite{davies2018loihi}. Given the extremely low idle energy consumption, the energy spent by SNNs for learning and inference is essentially proportional to the number of spikes processed and communicated by the neurons, with the energy per spike being as low as a few picojoules \cite{rajendran2019low}.

\textbf{Deterministic versus probabilistic SNN models.} 
The most common SNN model consists of a network of neurons with deterministic dynamics whereby a spike is emitted as soon as an internal state variable, known as the {\em membrane potential}, crosses a given threshold value. A typical example is the leaky integrate-and-fire model, in which the membrane potential increases with each spike recorded in the incoming synapses, while decreasing in the absence of inputs. When information is encoded in the rate of spiking of the neurons, an SNN can approximate the behavior of a conventional ANN with the same topology. This has motivated a popular line of work that aims at converting a pretrained ANN into a potentially more efficient SNN implementation (see \cite{rueckauer2018conversion} and the ``Models'' section for further details). 

To make full use of the temporal processing capabilities of SNNs, learning problems should be formulated as the minimization of a loss function that directly accounts for the timing of the spikes emitted by the neurons. As for ANNs, this minimization can, in principle, be done using stochastic gradient descent (SGD). Unlike ANNs, however, this conventional approach is made challenging by the nondifferentiability of the output of the SNN with respect to the synaptic weights due to the threshold crossing-triggered behavior of spiking neurons. The potentially complex recurrent topology of SNNs also makes it difficult to implement the standard backpropagation procedure used in multilayer ANNs to compute gradients. To obviate this problem, a number of existing learning rules approximate the derivative by smoothing out the membrane potential as a function of the weights \cite{lee2016training, OConnorW16,wu2018spatio}.

In contrast to deterministic models for SNNs, a probabilistic model defines the outputs of all spiking neurons as jointly distributed binary random processes. The joint distribution is differentiable in the synaptic weights, and, as a result, so are principled learning criteria from statistics and information theory, such as likelihood function and mutual information. The maximization of such criteria can apply to arbitrary topologies and does not require the implementation of backpropagation mechanisms. Hence, a stochastic viewpoint has significant analytic advantages, which translate into the derivation of flexible learning rules from first principles. These rules recover as special cases many known algorithms proposed for SNNs in the theoretical neuroscience literature as biologically plausible algorithms \cite{AbbotDayan}. 
 
\textbf{Scope and overview.} 
This article aims to provide a review on the topic of probabilistic SNNs with a specific focus on the most commonly used generalized linear models (GLMs). We cover models, learning rules, and applications, highlighting principles and tools. The main goal is to make key ideas in this emerging field accessible to researchers in signal processing, who may otherwise find it difficult to navigate the theoretical neuroscience literature on the subject, given its focus on biological plausibility rather than theoretical and algorithmic principles \cite{AbbotDayan}. At the end of the article, we also review alternative probabilistic formulations of SNNs, extensions, and open problems.

\section*{Learning Tasks} \label{sec:learn_task}

\begin{figure}[t!]
\centering
\subfigure[]{
\includegraphics[width=0.48\columnwidth]{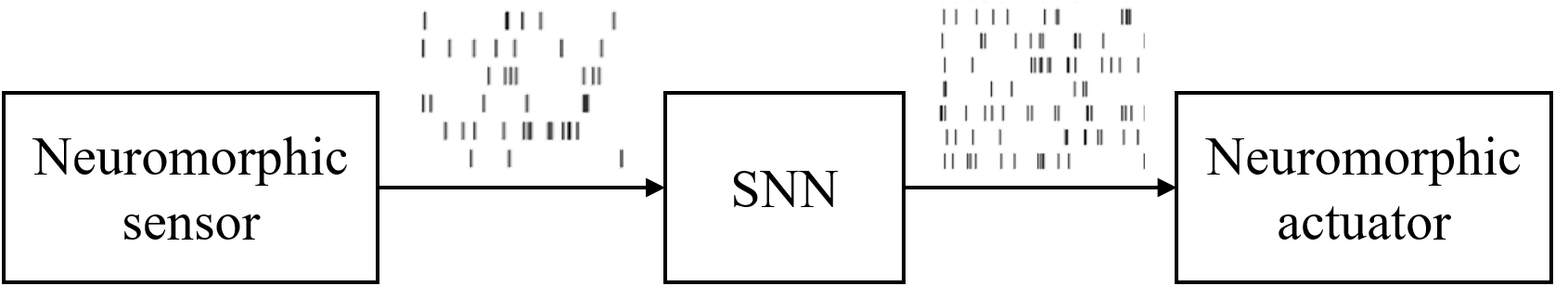} \label{fig:snn_io_a}
}
\subfigure[]{
\includegraphics[width=0.48\columnwidth]{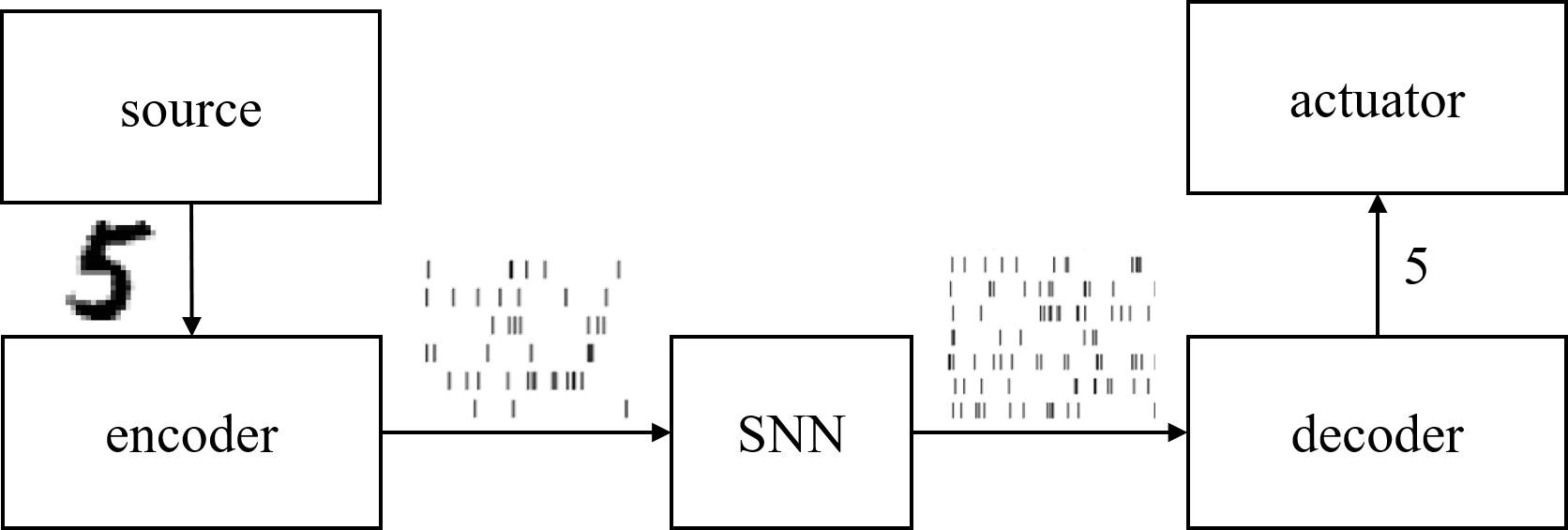} \label{fig:snn_io_b}
}
  \caption{Depictions of the input/output interfaces of an SNN: (a) a direct interface with a neuromorphic sensor and actuator and (b) an indirect interface through encoding and decoding.}
  \label{fig:snn_io}
  \vspace{-0.55cm}
\end{figure}

An SNN is a network of spiking neurons. As seen in Fig.~\ref{fig:snn_io}, the input and output interfaces of an SNN typically transfer spiking signals. Input spiking signals can either be recorded directly from neuromorphic sensors, such as silicon cochleas and retinas [Fig.~\ref{fig:snn_io_a}], or be converted from a natural signal to a set of spiking signals [Fig.~\ref{fig:snn_io_b}]. Conversion can be done by following different rules, including {\em rate encoding}, whereby amplitudes are converted into the (instantaneous) spiking rate of a neuron; {\em time encoding}, whereby amplitudes are translated into spike timings; and {\em population encoding}, whereby amplitudes are encoded into the (instantaneous) firing rates \cite{eliasmith2004neural} or relative firing times of a subset of neurons (see \cite{AbbotDayan} for a review). In a similar manner, output spiking signals can either be fed directly to a neuromorphic actuator, such as neuromorphic controllers or prosthetic systems [Fig.~\ref{fig:snn_io_a}], or be converted from spiking signals to natural signals [Fig.~\ref{fig:snn_io_b}]. This can be done by following rate, time, or population decoding principles.

The SNN generally acts as a dynamic mapping between inputs and outputs that is defined by the model parameters, including, most notably, the interneuron synaptic weights. This mapping can be designed or trained to carry out \emph{inference} or {\em control} tasks. When training is enabled, the model parameters are automatically adapted based on data fed to the network, with the goal of maximizing a given performance criterion. Training can be carried out in a supervised, unsupervised, or reinforcement learning manner, depending on the availability of data and feedback signals, as further discussed subsequently. For both inference/control and training, data can be presented to the SNN in a \emph{batch} mode (also known as frame-based mode) or in an \emph{online mode} (see the ``Training SNNs'' section).

With \emph{supervised learning}, the training data specify both input and desired output. Input and output pairs are either in the form of a number of separate examples, in the case of batch learning, or presented over time in a streaming fashion for online learning. As an example, the training set may include a number of spike-encoded images and corresponding correct labels, or a single time sequence to be used to extrapolate predictions (see also the ``Batch Learning Examples'' and ``Online Learning Examples'' sections). Under \emph{unsupervised learning}, the training data only specify only the desired input or output to the SNN, which can again be presented in a batch or online fashion. Examples of applications include representation learning, which aims to translate the input into a more compact, interpretable, or useful representation, and generative modeling, which seeks to generate outputs with statistics akin to the training data (see, e.g., \cite{simeone2018brief}). Finally, with \emph{reinforcement learning}, the SNN is used to control an agent on the basis of input observations from the environment to accomplish a given goal. To this end, the SNN is provided with feedback on the selected outputs that guides the SNN in updating its parameters in a batch or online manner \cite{sutton2018reinforcement}.


\section*{Models} \label{sec:model}

Here, we describe the standard discrete-time GLM for SNNs, also known as the {\em spike response model with escape noise} (see, e.g., \cite{pillow08:spatio} and \cite{gerstner2002spiking}). Discrete-time models reflect the operation of a number of neuromorphic chips, including Intel's Loihi \cite{davies2018loihi}, while continuous-time models are more commonly encountered in the computer neuroscience literature \cite{AbbotDayan}.

\begin{figure}[t!]
\centering
\subfigure[]{
\includegraphics[height=0.24\columnwidth]{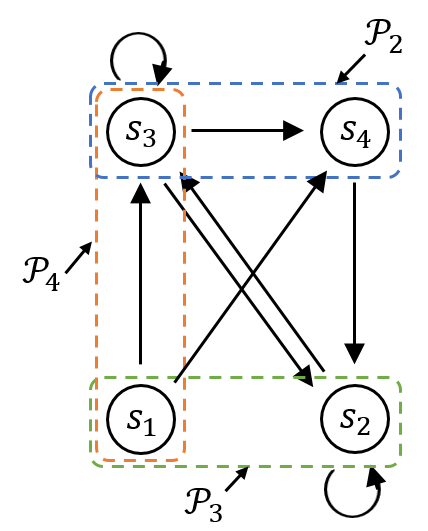}
}
\subfigure[]{
\includegraphics[height=0.24\columnwidth]{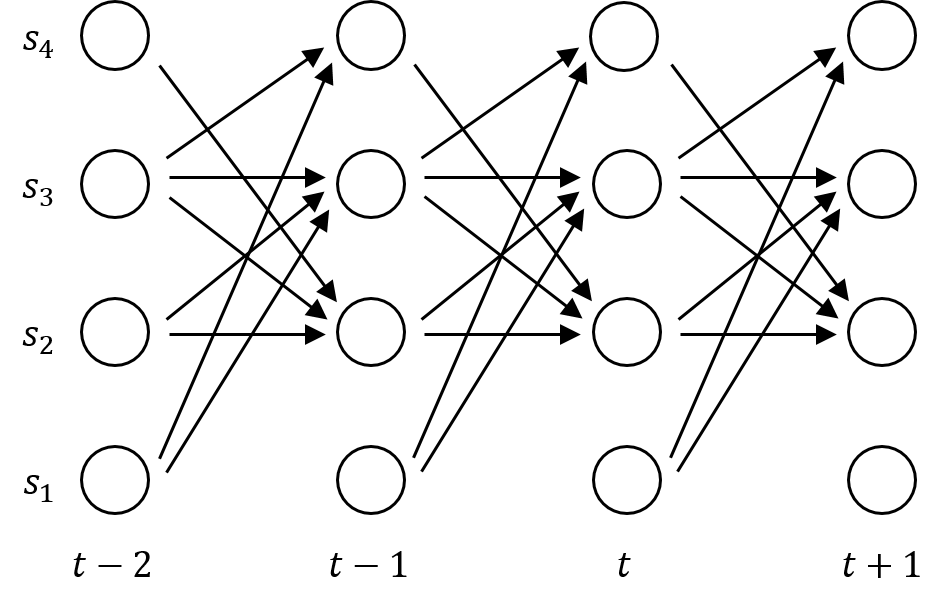}
}
  \caption{(a) An architecture of an SNN with $N = 4$ spiking neurons. The directed links between two neurons represent causal feedforward, or synaptic, dependencies, while the self-loop links represent feedback dependencies. The directed graph may have loops, including self-loops, indicating recurrent behavior. (b) A time-expanded view of the temporal dependencies implied by (a) with synaptic and feedback memories equal to one time step. }
  \label{fig:ex_model}
  \vspace{-0.55cm}
\end{figure}

\textbf{Graphical representation.} 
As illustrated in Fig.~\ref{fig:ex_model}, an SNN consists of a network of $N$ spiking neurons. At any time $t = 0,1,2,\ldots$, each neuron $i$ outputs a binary signal $s_{i,t} \in \{0,1\}$, with value $s_{i,t} = 1$ corresponding to a {\em spike} emitted at time $t$. We collect in vector $\bms_t = (s_{i,t} : i \in \set{V})$ the binary signals emitted by all neurons at time $t$, where $\set{V}$ is the set of all neurons. Each neuron $i \in \set{V}$ receives the signals emitted by a subset $\set{P}_i$ of neurons through directed links, known as {\em synapses}. Neurons in set $\set{P}_i$ are referred to as {\em presynaptic} for {\em postsynaptic} neuron $i$.

\begin{figure}[t!]
\centering
\includegraphics[height=0.33\columnwidth]{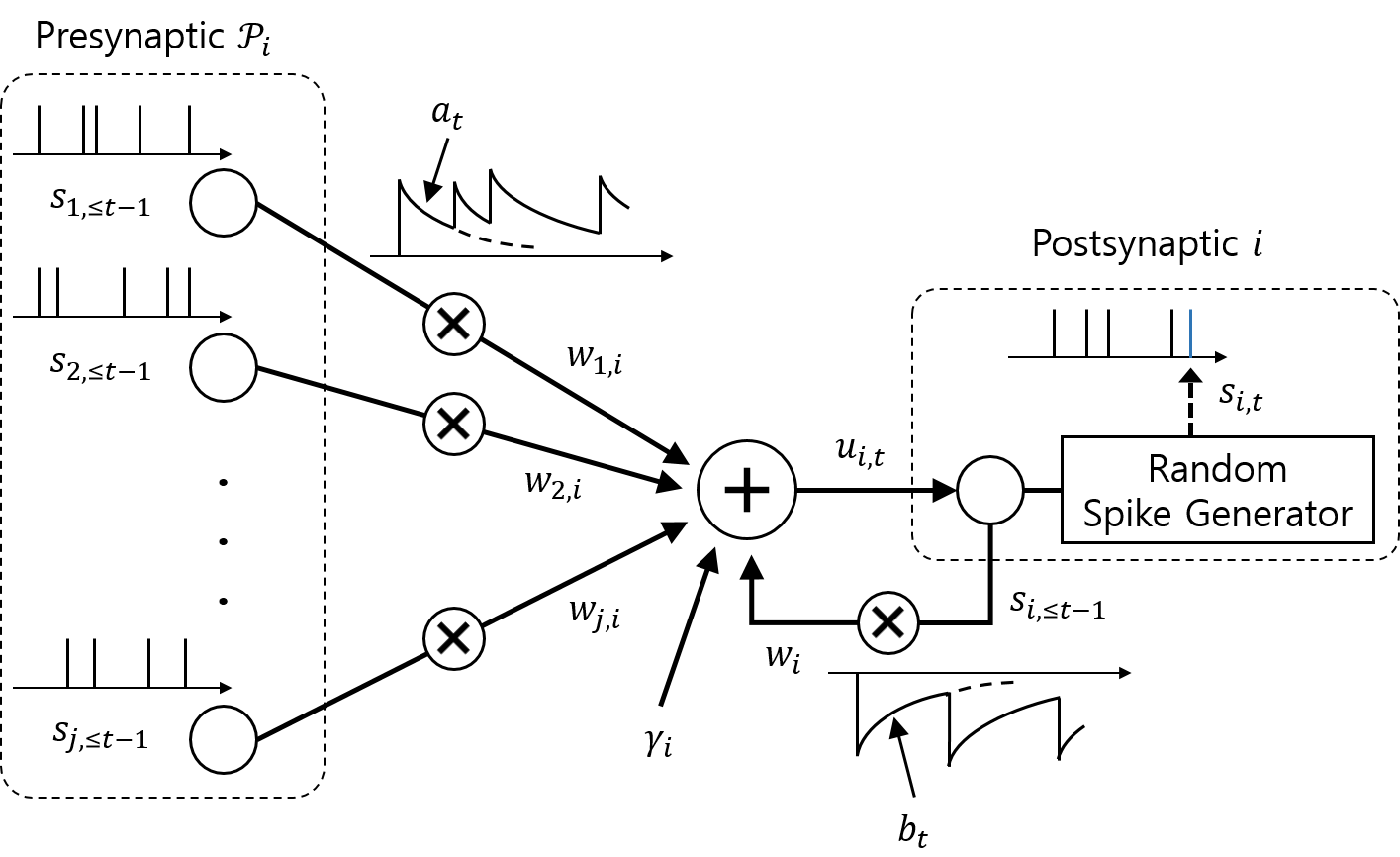}
\vspace{-0.2cm}
\caption{An illustration of the membrane potential model, with exponential feedforward and feedback kernels (see also Fig.~\ref{fig:ex_basis}). }
\label{fig:snn_LIF}
\vspace{-0.55cm}
\end{figure}

\textbf{Membrane potential and filtered traces.} 
The internal, analog state of each spiking neuron $i \in \set{V}$ at time $t$ is defined by its {\em membrane potential} $u_{i,t}$ (and possibly by other secondary variables to be discussed) \cite{gerstner2002spiking}. The value of the membrane potential indicates the probability of neuron $i$ to spike. As illustrated in Fig.~\ref{fig:snn_LIF}, the membrane potential is the sum of contributions from the incoming spikes of the presynaptic neurons and from the past spiking behavior of the neuron itself, where both contributions are filtered by the respective kernels $a_t$ and $b_t$. To elaborate, we denote as $\bms_{i,\leq t} = (s_{i,0},\ldots,s_{i,t})$ the spike signal emitted by neuron $i$ up to time $t$. Given past input spike signals from the presynaptic neurons $\set{P}_i$, denoted as $\bms_{\set{P}_i, \leq t-1} = \{ \bms_{j,\leq t-1} \}_{j \in \set{P}_i}$, and the local spiking history $\bms_{i, \leq t-1}$, the membrane potential of postsynaptic neuron $i$ at time $t$ can be written as \cite{gerstner2002spiking}
\begin{align} \label{eq:mem_potentiali_LIF}
u_{i,t} = \sum_{j \in \set{P}_i} w_{j,i} \overrs_{j,t-1} + w_i \overls_{i,t-1} + \gamma_i,
\end{align}
where the quantities $w_{j,i}$ for $j \in \set{P}_i$ are synaptic (feedforward) weights, $w_i$ is a feedback weight, $\gamma_i$ is a bias parameter, and the quantities
\begin{align} \label{eq:filtered_signal}
\overrs_{i,t} = a_t \ast s_{i,t} ~~\text{and}~~ \overls_{i,t} = b_t \ast s_{i,t}
\end{align}
are known as {\em filtered feedforward} and {\em feedback traces} of neuron $i$, respectively, where $\ast$ denotes the convolution operator $f_t \ast g_t = \sum_{\delta \geq 0} f_{\delta} g_{t-\delta}$.

\begin{figure}[t!]
\centering 
\subfigure[]{
\includegraphics[height=0.165\columnwidth]{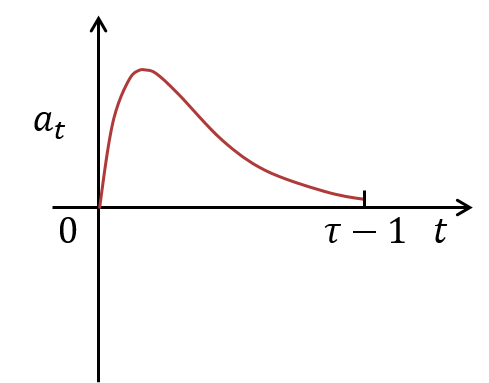} \label{fig:ex_basis_a}
} \hspace{-0.35cm}
\subfigure[]{
\includegraphics[height=0.165\columnwidth]{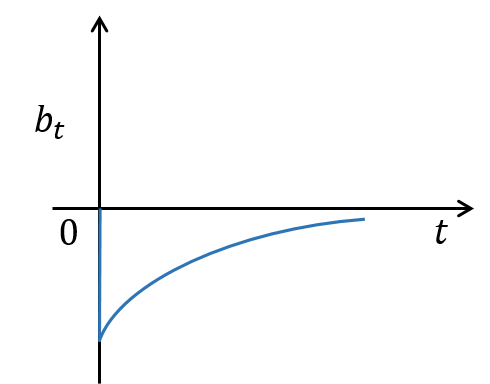} \label{fig:ex_basis_b}
} \hspace{-0.35cm}
\subfigure[]{
\includegraphics[height=0.165\columnwidth]{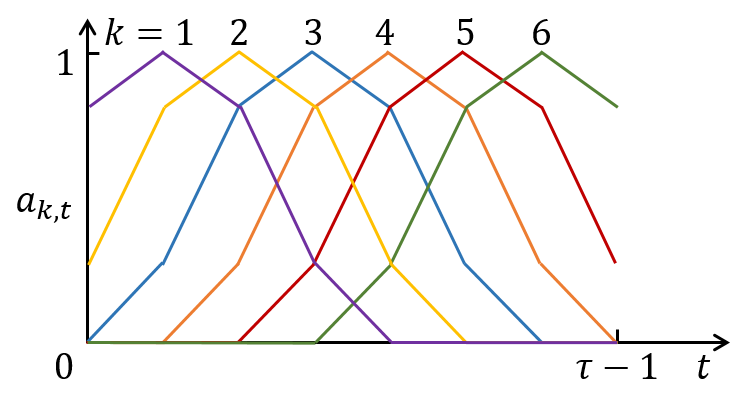} \label{fig:ex_basis_c}
} \hspace{-0.35cm}
\subfigure[]{
\includegraphics[height=0.165\columnwidth]{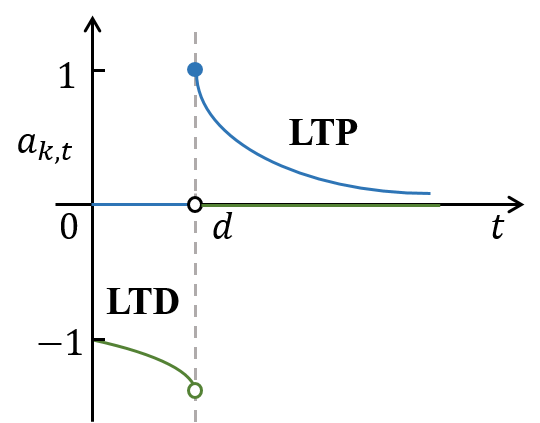} \label{fig:ex_basis_d}
}
  \caption{Examples of feedforward/feedback kernels: (a) an exponentially decaying feedforward kernel $a_t$, (b) an exponentially decaying feedback kernel $b_t$, (c) raised cosine basis functions $a_{k,t}$ in \cite{pillow08:spatio}, and (d) spike-timing-dependent plasticity basis functions $a_{k,t}$ for long-term potentiation (LTP) and long-term depression (LTD), where the synaptic conduction delay equals $d$ \cite{osogami17:BMtime}. }
  \label{fig:ex_basis}
  \vspace{-0.55cm}
\end{figure}

\textbf{Kernels and model weights.} In \eqref{eq:mem_potentiali_LIF}-\eqref{eq:filtered_signal}, the filter $a_t$ defines the synaptic response to a spike from a presynaptic neuron at the postsynaptic neuron. This filter is known as the {\em feedforward}, or {\em synaptic}, {\em kernel}. The filtered contribution of a spike from the presynaptic neuron $j \in \set{P}_i$ is multiplied by a learnable weight $w_{j,i}$ for the synapse from neuron $j$ to neuron $i \in \set{V}$. When the filter is of finite duration $\tau$, computing the feedforward trace $\overrs_{i,t}$ requires keeping track of the window $\{s_{i,t},s_{i,t-1},\ldots,s_{i,t-(\tau-1)} \}$ of prior synaptic inputs as part of the neuron's state \cite{osogami17:BMtime}. An example is given by the function $a_t = \big( \exp(-t/\tau_1) - \exp(-t/\tau_2) \big)$ for $t=0,...,\tau-1$ and zero otherwise, with time constants $\tau_1$ and $\tau_2$ and duration $\tau$, as illustrated in Fig.~\ref{fig:ex_basis_a}. When the kernel is chosen as an infinitely long decaying exponential, i.e., as $a_t = \exp (-t/\tau_1)$, the feedforward trace $\overrs_{i,t}$ can be directly computed using an autoregressive update that requires the storage of only a single scalar variable in the neuron's state \cite{osogami17:BMtime}, i.e., $\overrs_{i,t} = \exp(-1/\tau_1) (\overrs_{i,t-1} + s_{i,t})$. In general, the time constants and kernel shapes determine the synaptic memory and synaptic delays. 

The filter $b_t$ describes the response of a neuron to a local output spike and is known as a {\em feedback kernel}. A negative feedback kernel, such as $b_t = -\exp (-t/\tau_m)$, with time constant $\tau_m$ (see Fig.~\ref{fig:ex_basis_b}), models the refractory period upon the emission of a spike, with the time constant of the feedback kernel determining the duration of the refractory period. As per \eqref{eq:mem_potentiali_LIF}, the filtered contribution of a local output spike is weighted by a learnable parameter $w_i$. Similar considerations as for the feedforward traces apply regarding the computation of the feedback trace. 

Generalizing the model described previously, a synapse can be associated with $K_a$ learnable synaptic weights $\{ w_{j,i,k}\}_{k=1}^{K_a}$. In this case, the contribution from presynaptic neuron $j$ in \eqref{eq:mem_potentiali_LIF} can be written as \cite{pillow08:spatio} 
\begin{align} \label{eq:basis_feedforward}
\bigg( \sum_{k=1}^{K_a} w_{j,i.k} a_{k,t} \bigg) \ast s_{j,t},
\end{align}
where we have defined $K_a$ fixed basis functions $\{a_{k,t}\}_{k=1}^{K_a}$, with learnable weights $\{w_{j,i,k}\}_{k=1}^{K_a}$. The feedback kernel can be similarly parameterized as the weighted sum of fixed $K_b$ basis functions. Parameterization \eqref{eq:basis_feedforward} makes it possible to adapt the shape of the filter applied by the synapse by learning the weights $\{w_{j,i,k}\}_{k=1}^{K_a}$. Typical examples of basis functions are the raised cosine functions shown in Fig.~\ref{fig:ex_basis_c}. With this choice, the system can learn the sensitivity of each synapse to different synaptic delays, each corresponding to a different basis function, by adapting the weights $\{ w_{j,i,k}\}_{k=1}^{K_a}$. In the rest of this article, with the exception of the ``Batch Learning Examples'' and ``Online Learning Examples'' sections, we focus on the simpler model of \eqref{eq:mem_potentiali_LIF}-\eqref{eq:filtered_signal}.

Practical implementations of the membrane potential model \eqref{eq:mem_potentiali_LIF} can leverage the fact that linear filtering of binary spiking signals requires only carrying out sums while doing away with the need to compute expensive floating-point multiplications \cite{rajendran2019low}.

\textbf{GLM.} 
As discussed, a probabilistic model defines the joint probability distribution of the spike signals emitted by all neurons. In general, with the notation $\bms_{\leq t} = (\bms_0, \ldots, \bms_t)$ using the chain rule, the log-probability of the spike signals $\bms_{\leq T}= (\bms_0, \ldots, \bms_T)$ emitted by all neurons in the SNN up to time $T$ can be written as 
\begin{align} \label{eq:joint-time}
\log p_\bmtheta(\bms_{\leq T}) = \sum_{t=0}^T \log p_{\bmtheta}(\bms_t | \bms_{\leq t-1}) = \sum_{t=0}^T \sum_{i \in \set{V}} \log p_{\bmtheta_i}(s_{i,t} | \bms_{\set{P}_i \cup \{i\}, \leq t-1}),
\end{align}
where $\bmtheta = \{\bmtheta_i\}_{i \in \set{V}}$ is the learnable parameter vector, with $\bmtheta_i$ being the local parameters of neuron $i$. The decomposition \eqref{eq:joint-time} is in terms of the conditional probabilities $p_{\bmtheta_i}(s_{i,t} | \bms_{\set{P}_i \cup \{i\}, \leq t-1})$, which represent the spiking probability of neuron $i$ at time $t$, given its past spike timings and the past behaviors of its presynaptic neurons $\set{P}_i$. 

Under the GLM, the dependency of the spiking behavior of neuron $i \in \set{V}$ on the history $\bms_{\set{P}_i \cup \{i\}, \leq t-1}$ is mediated by the neuron's membrane potential $u_{i,t}$. Specifically, the instantaneous firing probability of neuron $i$ at time $t$ is equal to 
\begin{align} \label{eq:prob-ind}
p_{\bmtheta_i}(s_{i,t} = 1 | \bms_{\set{P}_i \cup \{i\}, \leq t-1}) = p(s_{i,t} = 1 | u_{i,t}) = \sigma(u_{i,t}),
\end{align}
with $\sigma(\cdot)$ being the sigmoid function, i.e., $\sigma(x) = 1/(1 + \exp(-x))$. According to \eqref{eq:prob-ind}, a larger potential $u_{i,t}$ increases the probability that neuron $i$ spikes. The model \eqref{eq:prob-ind} is parameterized by the local learnable vector $\bmtheta_i = \{ \gamma_i, \{ w_{j,i} \}_{j \in \set{P}_i}, w_i \}$ of neuron $i$. SNNs modeled according to the described GLM framework can be thought of as a generalization of dynamic models of belief networks \cite{neal1992connectionist}, and they can also be interpreted as a discrete-time version of Hawkes processes \cite{gerhard2017stability}.

In a variant of this model, probability \eqref{eq:prob-ind} can be written as $\sigma( u_{i,t}/\Delta u)$, where $\Delta u$ is a bandwidth parameter that dictates the smoothness of the firing rate about the threshold. When taking the limit $\Delta u \rightarrow 0$, we obtain the deterministic leaky integrate-and-fire model \cite{gardner2015learning}.

{\bf Relationship with ANNs.} 
Under rate encoding, as long as the duration $T$ is large enough, the deterministic integrate-and-fire model can mimic the operation of a conventional feedforward ANN with a nonnegative activation function. To this end, consider an ANN with an arbitrary topology defined by an \emph{acyclic} directed graph. The corresponding SNN has the same topology, a feedforward kernel defined by a single basis function implementing a perfect integrator (i.e., a filter with a constant impulse response), the same synaptic weights of the ANN, and no feedback kernel. In this way, the value of the filtered feedforward trace for each synapse approximates the spiking rate of the presynaptic neuron as $T$ increases. The challenge in enabling a conversion from ANN to SNN is to choose the thresholds $\gamma_i$ and possibly a renormalization of the weights, so that the spiking rates of all neurons in the SNN approximate the outputs of the neurons in the ANN \cite{rueckauer2018conversion}. When including loops, deterministic SNNs can also implement recurrent neural networks (RNNs) \cite{neftci2019surrogate}.

{\bf Gradient of the log-likelihood.} 
The gradient of the log probability, or {\em log-likelihood}, $\set{L}_{\bms_{\leq T}}(\bmtheta) = \log p_\bmtheta(\bms_{\leq T})$ in \eqref{eq:joint-time}, with respect to the learnable parameters $\bmtheta$, plays a key role in the problem of training a probabilistic SNN. Focusing on any neuron $i \in \set{V}$, from \eqref{eq:mem_potentiali_LIF} to \eqref{eq:prob-ind}, the gradient of the log-likelihood with respect to the local parameters $\bmtheta_i$ for neuron $i$ is given as
\begin{align} \label{eq:ll-grad-total}
\grad_{\bmtheta_i} \set{L}_{\bms_{\leq T}}(\bmtheta) = \sum_{t=0}^T \grad_{\bmtheta_i} \log p_{\bmtheta_i}(s_{i,t} | \bms_{\set{P}_i \cup \{i\}, \leq t-1}),
\end{align}
where the individual entries of the gradient of time $t$ can be obtained as
\begin{subequations} \label{eq:ll-grad}
\begin{align} 
\grad_{\gamma_i} \log p_{\bmtheta_i}(s_{i,t} | \bms_{\set{P}_i \cup \{i\}, \leq t-1}) &= 
s_{i,t} - \sigma(u_{i,t}), \label{eq:ll-grad-gamma} \\ 
\grad_{w_{j,i}} \log p_{\bmtheta_i}(s_{i,t} | \bms_{\set{P}_i \cup \{i\}, \leq t-1}) &= 
\overrs_{j,t-1} \Big( s_{i,t} - \sigma(u_{i,t}) \Big), \label{eq:ll-grad-feedforward}  \\
\text{and} \quad \grad_{w_{i}} \log p_{\bmtheta_i}(s_{i,t} | \bms_{\set{P}_i \cup \{i\}, \leq t-1}) &= 
\overls_{i,t-1} \Big( s_{i,t} - \sigma(u_{i,t}) \Big). \label{eq:ll-grad-feedback} 
\end{align}
\end{subequations}
The gradients \eqref{eq:ll-grad} depend on the difference between the desired spiking behavior and its average behavior under the model distribution \eqref{eq:prob-ind}. The implications of this result for learning will be discussed in the next sections.

\section*{Training SNNs} 
\label{sec:learning_snn}

SNNs can be trained using supervised, unsupervised, and reinforcement learning. To this end, the network follows a {\em learning rule}, which defines how the model parameters $\bmtheta$ are updated on the basis of the available observations. As we will detail, learning rules can be applied in a {\em batch} mode at the end of a full period $T$ of use of the SNN, based on multiple observations of duration $T$, or in an {\em online} fashion, i.e., after each time instant $t$, based on an arbitrarily long observation.

{\bf Locality.} 
A learning rule is {\em local} if its operation can be decomposed into atomic steps that can be carried out in parallel at distributed processors based only on locally available information and limited communication on the connectivity graph (see Fig.~\ref{fig:ex_model}). Local information at a neuron includes the membrane potential, the feedforward filtered traces for the incoming synapses, the local feedback filtered trace, and the local model parameters. The processors will be considered here to be conventionally implemented at the level of individual neurons. Beside local signals, learning rules may also require global feedback signals, as discussed next.

{\bf Three-factor rule.} 
While the details differ for each learning rule and task, a general form of the learning rule for the synaptic weights follows the {\em three-factor rule} \cite{fremaux2016neuromodulated, brea2013matching}. Accordingly, the synaptic weight $w_{j,i}$ from presynaptic neuron $j \in \set{P}_i$ to a postsynaptic neuron $i \in \set{V}$ is updated as 
\begin{align} \label{eq:three}
w_{j,i} &\leftarrow w_{j,i} + \eta \times \ell \times \text{pre}_j \times \text{post}_i, 
\end{align}
where $\eta$ is a learning rate, $\ell$ is a scalar {\em global learning signal} that determines the sign and magnitude of the update, $\text{pre}_j$ is a function of the activity of the presynaptic neuron $j \in \set{P}_i$, and $\text{post}_i$ depends on the activity of the postsynaptic neuron $i \in \set{V}$. For most learning rules, pre- and postsynaptic terms are local to each neuron, while the learning signal $\ell$, if present, plays the role of a global feedback signal. As a special case, the rule \eqref{eq:three} can implement Hebb's hypothesis that ``neurons that spike together wire together''. This is indeed the case if the product of $\text{pre}_j$ and $\text{post}_i$ terms is large when the two neurons spike at nearly the same time, resulting in a large change of the synaptic weight $w_{j,i}$ \cite{AbbotDayan}. 

In the next two sections, we will see how learning rules of the form \eqref{eq:three} can be derived in a principled manner as SGD updates obtained under the described probabilistic SNN models.

\section*{Training SNNs: Fully observed models} 
\label{sec:learning_prob_fully}

\textbf{Fully observed vs partially observed models.} 
Neurons in an SNN can be divided into the subset of {\em visible}, or {\em observed}, neurons, which encode inputs and outputs, and {\em hidden}, or {\em latent}, neurons, whose role is to facilitate the desired behavior of the SNN. During training, the behavior of visible neurons is specified by the training data. For example, under supervised learning, input neurons are clamped to the input data, while the spiking signals of output neurons are determined by the desired output. Another related example is a reinforcement learning task in which the SNN models a policy, with input neurons encoding the state and output neurons encoding the action previously taken by the learner in response to the given input \cite{bleema2018fts}.

In the case of {\em fully observed models}, the SNN contains only visible neurons while, in the case of {\em partially observed models}, the SNN also includes hidden neurons. We first consider the simpler former case and then extend the discussion to partially observed models.  

\vspace{-0.4cm}
\subsection*{Maximum likelihood learning via SGD} 

The standard training criterion for probabilistic models for both supervised and unsupervised learning is maximum likelihood (ML). ML selects model parameters that maximize the probability of the observed data and, hence, of the desired input/output behavior under the model. To elaborate, we consider an example $\bmx_{\leq T}$ consisting of fully observed spike signals for all neurons in the SNN, including both input and output neurons. Using the notation in the ``Models'' section, we hence have $\bms_{\leq T} = \bmx_{\leq T}$. During training, the spike signals for all neurons are thus clamped to the values assumed in the data point $\bmx_{\leq T}$, and the log-likelihood is given as $\set{L}_{\bmx_{\leq T}}(\bmtheta) = \log p_\bmtheta(\bmx_{\leq T})$ in \eqref{eq:joint-time}, with $\bms_{\leq T} = \bmx_{\leq T}$.	As we will see next, for batch learning, there are multiple such examples $\bmx_{\leq T}$ in the training set while, for online learning, we have a single arbitrary long example $\bmx_{\leq T}$ for large $T$.

{\bf Batch SGD.} 
In the batch training mode, a training set $\set{D} = \{\bmx_{\leq T}^m \}_{m=1}^M$ of $M$ fully observed examples is available to enable the learning of the model parameters. The batch SGD-based rule proceeds iteratively by selecting an example $\bmx_{\leq T}$ from the training set $\set{D}$ at each iteration (see, e.g., \cite{goodfellow2016deep}). The model parameters $\bmtheta$ are then updated in the direction of the gradient \eqref{eq:ll-grad-total} and \eqref{eq:ll-grad}, with $\bms_{\leq T} = \bmx_{\leq T}$, as
\begin{align} \label{eq:fully-ml-sgd-batch}
\bmtheta \leftarrow \bmtheta + \eta \grad_\bmtheta \set{L}_{\bmx_{\leq T}}(\bmtheta),
\end{align}
where the learning rate $\eta$ is assumed to be fixed here for simplicity of notation. Note that the update \eqref{eq:fully-ml-sgd-batch} is applied at the end of the observation period $T$. The batch algorithm can be generalized by summing over a minibatch of examples at each iteration \cite{goodfellow2016deep}. 

{\bf Online SGD.} 
In the online training mode, an arbitrary long example $\bmx_{\leq T}$ is available, and the model parameters $\bmtheta$ are updated at each time $t$ (or, more generally, periodically every few time instants). This can be done by introducing an {\em eligibility trace} $\bme_{i,t}$ for each neuron $i$ \cite{gardner2015learning, brea2013matching}. As summarized in Algorithm~\ref{alg:fully-ml-sgd}, the eligibility trace $\bme_{i,t}$ in \eqref{eq:eligibility}, with $\kappa < 1$, computes a weighted average of current and past gradient updates. In this update, the current gradient $\grad_\bmtheta \log p_\bmtheta(\bmx_t | \bmx_{\leq t-1})$ is weighted by a factor $(1-\kappa)$, and the gradient that is evaluated $l$ steps in the past is multiplied by the exponentially decaying coefficient $(1-\kappa) \cdot \kappa^l$. The eligibility trace captures the impact of past updates on the current spiking behavior, and it can help stabilize online training by reducing the variance of the updates (for sufficiently large $\kappa$) \cite{sutton2018reinforcement}.

\begin{algorithm}[t]
\setstretch{0.975}
\DontPrintSemicolon
\LinesNumbered
\KwIn{Training example $\bmx_{\leq T}$ and learning rates $\eta$ and $\kappa$}
\KwOut{Learned model parameters $\bmtheta$}
\vspace{0.1cm}
\hrule
\vspace{0.1cm}
{\bf initialize} parameters $\bmtheta$ \\
\Repeat{{\em stopping criterion is satisfied}}{
for each $t=0,1,\ldots,T$ \\
\For{{\em each neuron $i \in \set{V}$ }}{
compute the gradient $\grad_{\bmtheta_i} \log p_{\bmtheta_i}(x_{i,t} | \bmx_{\set{P}_i \cup \{i\}, \leq t-1})$ with respect to the local parameters $\bmtheta_i$ from \eqref{eq:ll-grad} \\
compute the eligibility trace $\bme_{i,t}$ 
\begin{align} \label{eq:eligibility}
\begin{aalign}  
\bme_{i,t} = \kappa \bme_{i,t-1} + (1-\kappa) \grad_{\bmtheta_i} \log p_{\bmtheta_i}(x_{i,t} | \bmx_{\set{P}_i \cup \{i\}, \leq t-1})    
\end{aalign}
\end{align} \\
update the local model parameters
\begin{align} \label{eq:fully-ml-sgd-online}
\begin{aalign}
\bmtheta_i \leftarrow \bmtheta_i + \eta \bme_{i,t}
\end{aalign}
\end{align}
}
\vspace{0.05cm}
} 
\caption{ML Training via online SGD}
\label{alg:fully-ml-sgd}
\end{algorithm}

{\bf Interpretation.} 
The online gradient update for any synaptic weight $w_{j,i}$ can be interpreted in light of the general form of rule \eqref{eq:three}. In fact, the gradient \eqref{eq:ll-grad-feedforward} has a {\em two-factor} form, whereby the global learning signal is absent; the presynaptic term is given by the filtered feedforward trace $\overrx_{j,t-1}$ of the presynaptic neuron $j \in \set{P}_i$, and the postsynaptic term is given by the error term $x_{i,t} - \sigma(u_{i,t})$. This error measures the difference between the desired spiking behavior of the postsynaptic neuron $i$ at any time $t$ and its average behavior under the model distribution \eqref{eq:prob-ind}. 

This update can be related to the standard spike-timing-dependent plasticity (STDP) rule \cite{AbbotDayan, osogami17:BMtime, bienenstock1982bcm}. In fact, STDP stipulates that the long-term potentiation (LTP) of a synapse occurs when the presynaptic neuron spikes right before a postsynaptic neuron, while long-term depression (LTD) of a synapse takes place when the presynaptic neuron spikes right after a postsynaptic neuron. With the basis functions depicted in Fig.~\ref{fig:ex_basis_d}, if a presynaptic spike occurs more than $d$ steps prior to the postsynaptic spike at time $t$, an increase in the synaptic weight, or LTP, occurs, while a decrease in the synaptic weight, or LTD, takes place otherwise \cite{osogami17:BMtime}. The parameter $d$ can hence be interpreted as synaptic delay.

As for the synaptic weights, all other gradients \eqref{eq:ll-grad} also depend on an error signal measuring the gap between the desired and average model behavior. In \eqref{eq:ll-grad-gamma}-\eqref{eq:ll-grad-feedback}, the desired behavior is given by samples $s_{i,t} = x_{i,t}$ in the training example. The contribution of this error signal can be interpreted as a form of (task-specific) {\em homeostatic plasticity}, in that it regulates the neuronal firing rates around desirable set-point values \cite{watt10:homeostatic, AbbotDayan}.

{\bf Locality and implementation.} 
Given the absence of a global learning signal, the online SGD rule in Algorithm~\ref{alg:fully-ml-sgd} and the batch SGD rule can be implemented locally, so that each neuron $i$ updates its own local parameters $\bmtheta_i$. Each neuron $i$ uses information about the local spike signal $x_{i,t}$, the feedforward filtered traces $\overrx_{j,t-1}$ for all presynaptic neurons $j \in \set{P}_i$, and the local feedback filtered trace $\overlx_{i,t-1}$ to compute the first terms in \eqref{eq:ll-grad-gamma}-\eqref{eq:ll-grad-feedback}, while the second terms in \eqref{eq:ll-grad-gamma}-\eqref{eq:ll-grad-feedback} are obtained from \eqref{eq:prob-ind} by using the neuron's membrane potential $u_{i,t}$.

\section*{Training SNNs: Partially observed models} 
\label{sec:learning_prob_latent}

{\bf Latent neurons.} 
As mentioned previously, the set $\set{V}$ of neurons can be partitioned into the disjoint subsets of observed (input and output) and hidden neurons. The $N_\textX$ neurons in the subset $\set{X}$ are observed, and the $N_\textH$ neurons in the subset $\set{H}$ are hidden, or latent, and we have $\set{V} = \set{X} \cup \set{H}$. We write as $\bmx_t = (x_{i,t}: i \in \set{X})$ and $\bmh_t = (h_{i,t}: i \in \set{H})$ the binary signals emitted by the observed and hidden neurons at time $t$, respectively. Therefore, using the notation in the ``Models'' section, we have $s_{i,t} = x_{i,t}$ for any observed neuron $i \in \set{X}$ and $s_{i,t} = h_{i,t}$ for any latent neuron $i \in \set{H}$ as well as $\bms_t = (\bmx_t, \bmh_t)$ for the overall set of spike signals at time $t$. During training, the spike signals $\bmx_{\leq T}$ of the observed neurons are clamped to the examples in the training set while the probability distribution of the signals $\bmh_{\leq T}$ of the hidden neurons can be adapted to ensure the desired input/output behavior. Mathematically, the probabilistic model is defined as in \eqref{eq:joint-time} and \eqref{eq:prob-ind}, with $\bms_{\leq T} = (\bmx_{\leq T}, \bmh_{\leq T})$.

\vspace{-0.4cm}
\subsection*{ML via SGD and variational learning}

Here, we review a standard learning rule that tackles the ML problem by using SGD. Unlike in the fully observed case, as we will see, {\em variational inference} is needed to cope with the complexity of computing the gradient of the log-likelihood of the observed spike signals in the presence of hidden neurons \cite{simeone2018brief}.

{\bf Log-likelihood.} 
The log-likelihood of an example of observed spike signals $\bmx_{\leq T}$ is obtained via marginalization by summing over all possible values of the latent spike signals $\bmh_{\leq T}$ as $\set{L}_{\bmx_{\leq T}}(\bmtheta) = \log p_\bmtheta(\bmx_{\leq T}) = \log \sum_{\bmh_{\leq T}} p_\bmtheta(\bmx_{\leq T}, \bmh_{\leq T})$. Let us denote as $\langle \cdot \rangle_p$ the expectation over a distribution $p$, as in $\langle f(x) \rangle_{p(x)} = \sum_{x} f(x)p(x)$, for some function $f(x)$. The gradient of the log-likelihood with respect to the model parameters $\bmtheta$ can be expressed as (see, e.g., \cite[Ch. 6]{simeone2018brief}) 
\begin{align} \label{eq:ll-latent-exact}
\grad_\bmtheta \set{L}_{\bmx_{\leq T}}(\bmtheta) = \Big\langle \grad_\bmtheta \log p_\bmtheta(\bmx_{\leq T}, \bmh_{\leq T}) \Big\rangle_{p_\bmtheta(\bmh_{\leq T} | \bmx_{\leq T})}, 
\end{align}
where the expectation is with respect to the posterior distribution $p_\bmtheta(\bmh_{\leq T} | \bmx_{\leq T})$ of the latent variables $\bmh_{\leq T}$, given the observation $\bmx_{\leq T}$. Note that the gradient $$\grad_\bmtheta \log p_\bmtheta (\bmx_{\leq T}, \bmh_{\leq T}) = \sum_{t=0}^T \grad_\bmtheta \log p_\bmtheta (\bmx_t, \bmh_t | \bmx_{\leq t-1}, \bmh_{\leq t-1})$$ is obtained from \eqref{eq:ll-grad}, with $\bms_{\leq T} = (\bmx_{\leq T}, \bmh_{\leq T})$. Computing the posterior $p_\bmtheta(\bmh_{\leq T} | \bmx_{\leq T})$ amounts to the Bayesian inference of the hidden spike signals for the observed values $\bmx_{\leq T}$. Given that we have the equality $p_\bmtheta(\bmh_{\leq T} | \bmx_{\leq T}) = p_\bmtheta(\bmx_{\leq T}, \bmh_{\leq T})/p_\bmtheta(\bmx_{\leq T})$, this task requires the evaluation of the marginal distribution $p_\bmtheta(\bmx_{\leq T}) = \sum_{\bmh_{\leq T}} p_\bmtheta(\bmx_{\leq T}, \bmh_{\leq T})$. For problems of practical size, this computation is intractable and, hence, so is evaluating the gradient \eqref{eq:ll-latent-exact}.

{\bf Variational learning.}
Variational inference, or variational Bayes, approximates the true posterior distribution $p_\bmtheta(\bmh_{\leq T} | \bmx_{\leq T})$ by means of any arbitrary {\em variational posterior} distribution $q_\bmphi(\bmh_{\leq T} | \bmx_{\leq T})$ parameterized by a vector $\bmphi$ of learnable parameters. For any variational distribution $q_\bmphi(\bmh_{\leq T} | \bmx_{\leq T})$, using Jensen's inequality, the log-likelihood $\set{L}_{\bmx_{\leq T}}(\bmtheta)$ can be lower bounded as (see, e.g., \cite[Ch. 6 and Ch. 8]{simeone2018brief})
\begin{align} \label{eq:elbo-general}
\set{L}_{\bmx_{\leq T}}(\bmtheta) = \log \sum_{\bmh_{\leq T}} p_\bmtheta(\bmx_{\leq T}, \bmh_{\leq T}) 
&\geq \sum_{\bmh_{\leq T}} q_\bmphi(\bmh_{\leq T} | \bmx_{\leq T}) \log \frac{p_\bmtheta(\bmx_{\leq T}, \bmh_{\leq T})}{q_\bmphi(\bmh_{\leq T}| \bmx_{\leq T})} \cr 
&= \Big\langle \ell_{\bmtheta, \bmphi} \big( \bmx_{\leq T}, \bmh_{\leq T} \big) \Big\rangle_{q_\bmphi(\bmh_{\leq T} | \bmx_{\leq T})} := L_{\bmx_{\leq T}}(\bmtheta, \bmphi),
\end{align}
where we have defined the {\em learning signal} as
\begin{align} \label{eq:ls-general}
\ell_{\bmtheta, \bmphi}(\bmx_{\leq T}, \bmh_{\leq T}) := \log p_\bmtheta(\bmx_{\leq T}, \bmh_{\leq T}) - \log q_\bmphi(\bmh_{\leq T} | \bmx_{\leq T}).
\end{align}

A baseline variational learning rule, also known as the {\em variational expectation maximization} algorithm, is based on the maximization of the evidence lower bound (ELBO) $L_{\bmx_{\leq T}}(\bmtheta, \bmphi)$ in \eqref{eq:elbo-general} with respect to both the model parameters $\bmtheta$ and the variational parameters $\bmphi$. Accordingly, for a given observed example $\bmx_{\leq T} \in \set{D}$, the learning rule is given by gradient ascent updates, where the gradients can be computed as
\begin{subequations} \label{eq:ll-elbo-general}
\begin{align} 
&\grad_\bmtheta L_{\bmx_{\leq T}} (\bmtheta, \bmphi) = \Big\langle \grad_\bmtheta \log p_{\bmtheta}(\bmx_{\leq T}, \bmh_{\leq T}) \Big\rangle_{q_\bmphi(\bmh_{\leq T} | \bmx_{\leq T})}, ~\text{and}~ \label{eq:ll-elbo-model} \\ 
&\grad_\bmphi L_{\bmx_{\leq T}} (\bmtheta, \bmphi) = \Big\langle \ell_{\bmtheta, \bmphi}(\bmx_{\leq T}, \bmh_{\leq T}) \cdot \grad_\bmphi \log q_\bmphi(\bmh_{\leq T} | \bmx_{\leq T}) \Big\rangle_{q_\bmphi(\bmh_{\leq T} | \bmx_{\leq T})}, \label{eq:ll-elbo-var}
\end{align}
\end{subequations}
respectively. The gradient \eqref{eq:ll-elbo-model} is derived in a manner analogous to \eqref{eq:ll-latent-exact}, and the gradient \eqref{eq:ll-elbo-var} is obtained from the standard REINFORCE, or score function, gradient \cite[Ch. 8]{simeone2018brief}, \cite{mnih14:nvil}. Importantly, the gradients \eqref{eq:ll-elbo-general} require expectations with respect to the known variational posterior $q_\bmphi(\bmh_{\leq T} | \bmx_{\leq T})$ evaluated at the current value of variational parameters $\bmphi$ rather than with respect to the hard-to-compute posterior $p_\bmtheta(\bmh_{\leq T} | \bmx_{\leq T})$. An alternative to the computation of the gradient over the variational parameters $\bmphi$ as in \eqref{eq:ll-elbo-var} is given by the so-called reparameterization trick \cite{kingma14:vae}, as briefly discussed in the ``Conclusions and Open Problems'' section.

In practice, computing the averages in \eqref{eq:ll-elbo-general} is still intractable because of the large domain of the hidden variables $\bmh_{\leq T}$. Therefore, the expectations over the variational posterior are typically approximated by means of Monte Carlo empirical averages. This is possible as long as sampling from the variational posterior $q_\bmphi(\bmh_{\leq T} | \bmx_{\leq T})$ is feasible. As an example, if a single spike signal ${\bmh}_{\leq T}$ is sampled from $q_\bmphi(\bmh_{\leq T} | \bmx_{\leq T})$, we obtain the Monte Carlo approximations of \eqref{eq:ll-elbo-general} as 
\begin{subequations} \label{eq:ll-elbo-mc}
\begin{align}
\grad_\bmtheta \hat{L}_{\bmx_{\leq T}}(\bmtheta,\bmphi) &= \grad_\bmtheta \log p_\bmtheta(\bmx_{\leq T}, {\bmh}_{\leq T}), ~\text{and}~ \label{eq:ll-elbo-mc-model}  \\ 
\grad_\bmphi \hat{L}_{\bmx_{\leq T}}(\bmtheta,\bmphi) &= \ell_{\bmtheta,\bmphi}(\bmx_{\leq T}, {\bmh}_{\leq T}) \cdot \grad_\bmphi \log q_\bmphi({\bmh}_{\leq T} | \bmx_{\leq T}). \label{eq:ll-elbo-mc-var}
\end{align}
\end{subequations}

{\bf Batch doubly SGD.} 
In a batch training formulation, at each iteration, an example $\bmx_{\leq T}$ is selected from the training set $\set{D}$. At the end of the observation period $T$, both model and variational parameters can be updated in the direction of the gradients $\grad_\bmtheta \hat{L}_{\bmx_{\leq T}}(\bmtheta,\bmphi)$ and $\grad_\bmphi \hat{L}_{\bmx_{\leq T}}(\bmtheta,\bmphi)$ in \eqref{eq:ll-elbo-mc} as 
\begin{subequations} \label{eq:latent-ml-vl-general}
\begin{align}
\bmtheta &\leftarrow \bmtheta + \eta_\bmtheta \grad_{\bmtheta} \hat{L}_{\bmx_{\leq T}}(\bmtheta, \bmphi), ~\text{and}~ \label{eq:latent-ml-vl-general-model} \\
\bmphi &\leftarrow \bmphi + \eta_\bmphi \grad_{\bmphi} \hat{L}_{\bmx_{\leq T}}(\bmtheta, \bmphi), \label{eq:latent-ml-vl-general-var}
\end{align}
\end{subequations}
respectively, where the learning rates $\eta_\bmtheta$ and $\eta_\bmphi$ are assumed to be fixed for simplicity. Rule \eqref{eq:latent-ml-vl-general} is known as {\em doubly SGD} since sampling is carried out over both the observed examples $\bmx_{\leq T}$ in the training set and the hidden spike signals $\bmh_{\leq T}$.

The doubly stochastic gradient estimator \eqref{eq:ll-elbo-mc-var} typically exhibits a high variance. To reduce the variance, a common approach is to subtract a {\em baseline control variate} from the learning signal. This can be done by replacing the learning signal in \eqref{eq:ll-elbo-mc-var} with the centered learning signal $\ell_{\bmtheta, \bmphi}(\bmx_{\leq T}, {\bmh}_{\leq T}) - \bar{\ell}$, where the baseline $\bar{\ell}$ is calculated as a moving average of learning signals computed at previous iterations \cite{mnih14:nvil, brea2013matching, rezende14:vi_snn}.

{\bf Online doubly SGD.} 
The batch doubly SGD rule \eqref{eq:latent-ml-vl-general} applies with any choice of variational distribution $q_\bmphi(\bmh_{\leq T} | \bmx_{\leq T})$, as long as it is feasible to sample from it and to compute the gradient in \eqref{eq:ll-elbo-mc-var}. However, the locality properties and complexity of the learning rule are strongly dependent on the choice of the variational distribution. We now discuss a specific choice considered in \cite{osogami17:BMtime,brea2013matching}, and \cite{rezende14:vi_snn,hinton00:sBM,kappel15:synaptic} that yields an online rule, summarized in Algorithm~\ref{alg:latent-ml-vl}.

The approach approximates the true posterior $p_\bmtheta(\bmh_{\leq T} | \bmx_{\leq T})$ with a feedforward distribution that ignores the stochastic dependence of the hidden spike signals $\bmh_t$ at time $t$ on the future values of the observed spike signals $\bmx_{\leq T}$. The corresponding variational distribution can be written as
\begin{align} \label{eq:var-posterior}
q_{\bmtheta^\textH}(\bmh_{\leq T} | \bmx_{\leq T}) = \prod_{t=0}^T p_{\bmtheta^\textH} (\bmh_t | \bmx_{\leq t-1}, \bmh_{\leq t-1}) = \prod_{t=0}^T \prod_{i \in \set{H}} p(h_{i,t} | u_{i,t}),
\end{align}
where we denote as $\bmtheta^\textH = \{\bmtheta_i\}_{i \in \set{H}}$ the collection of the model parameters for hidden neurons, and $p(h_{i,t}=1 | u_{i,t}) = \sigma(u_{i,t})$ by \eqref{eq:prob-ind}, with $s_{i,t} = h_{i,t}$. We note that \eqref{eq:var-posterior} is an approximation of the true posterior $p_\bmtheta(\bmh_{\leq T} | \bmx_{\leq T}) = \prod_{t=0}^T p_\bmtheta (\bmh_t | \bmx_{\leq T}, \bmh_{\leq t-1})$ since it neglects the correlation between variables $\bmh_t$ and the future observed samples $\bmx_{\geq t}$. In \eqref{eq:var-posterior}, we have emphasized that the variational parameters $\bmphi$ are tied to a subset of the model parameters, as per the equality $\bmphi = \bmtheta^\textH$. As a result, this choice of variational distribution does not include additional learnable parameters apart from the model parameters $\bmtheta$. The learning signal \eqref{eq:ls-general} with the feedforward distribution \eqref{eq:var-posterior} reads 
\begin{align} \label{eq:ls-var}
 \ell_{\bmtheta^\textX}(\bmx_{\leq T}, \bmh_{\leq T}) &= \sum_{t=0}^T \log p_{\bmtheta^\textX}(\bmx_t | \bmx_{\leq t-1}, {\bmh}_{\leq t-1}) = \sum_{t=0}^T \sum_{i \in \set{X}} \log p(x_{i,t} | u_{i,t}), 
\end{align}
where $\bmtheta^\textX = \{ \bmtheta_i\}_{i \in \set{X}}$ is the collection of the model parameters for observed neurons.

\begin{algorithm}[t]
\DontPrintSemicolon
\LinesNumbered
\vspace{0.05cm}
\KwIn{Training data $\bmx_{\leq T}$ and learning rates $\eta$ and $\kappa$}
\KwOut{Learned model parameters $\bmtheta$}
\vspace{0.1cm}
\hrule
\vspace{0.1cm}
{\bf initialize} parameters $\bmtheta$ \\
\Repeat{{\em stopping criterion is satisfied}}{
{\bf \em feedforward sampling:} \\
\For{{\em each hidden neuron $i \in \set{H}$}}{
emit a spike ${h}_{i,t} = 1$ with probability $\sigma({u}_{i,t})$ \\
}
{\bf \em global feedback:} \\
a central processor collects the log probabilities $p(x_{i,t} | {u}_{i,t})$ in \eqref{eq:prob-ind} from all observed neurons $i \in \set{X}$, computes a time-averaged learning signal \eqref{eq:ls-var} as
\begin{align} \label{eq:eligibility-ls}
\begin{aalign}
{\ell}_{t} = \kappa {\ell}_{t-1} + (1-\kappa) \sum_{i \in \set{X}} \log p(x_{i,t} | {u}_{i,t}),
\end{aalign}
\end{align}
and feeds back the global learning signal ${\ell}_t$ to all latent neurons \\
{\bf \em parameter update:} \\
\For{{\em each neuron $i \in \set{V}$}}{
evaluate the eligibility trace $\bme_{i,t}$ as 
\begin{align} \label{eq:eligibility-grad-var}
\begin{aalign}
\bme_{i,t} = 
\kappa \bme_{i,t-1} + (1-\kappa) \grad_{\bmtheta_i} \log p_{\bmtheta_i}(s_{i,t} | \bmx_{\leq t-1} \bmh_{\leq t-1}),
\end{aalign}
\end{align}
with $s_{i,t} = x_{i,t}$ if $i \in \set{X}$ and $s_{i,t} = h_{i,t}$ if $i \in \set{H}$ \\
update the local model parameters as 
\begin{align} \label{eq:latent-vl-sgd-online}
\begin{aalign}
\bmtheta_i \leftarrow \bmtheta_i + \eta \cdot
\begin{cases} 
\bme_{i,t}, \quad &\text{if}~ i \in \set{X} \\
\ell_t \bme_{i,t}, \quad &\text{if}~ i \in \set{H}
\end{cases}
\end{aalign}
\end{align}
}
}
\caption{ML Training via online Doubly SGD}
\label{alg:latent-ml-vl}
\end{algorithm}

With the choice \eqref{eq:var-posterior} for the variational posterior, the batch doubly SGD update rule \eqref{eq:latent-ml-vl-general} can be turned into an online rule by generalizing Algorithm~\ref{alg:fully-ml-sgd}, as detailed in Algorithm~\ref{alg:latent-ml-vl}. At each step of the online procedure, each hidden neuron $i \in \set{H}$ emits a spike, i.e., ${h}_{i,t} = 1$, at any time $t$ by following the current model distribution \eqref{eq:var-posterior}, i.e., with probability $\sigma({u}_{i,t})$. Note that the membrane potential ${u}_{i,t}$ of any neuron $i$ at time $t$ is obtained from \eqref{eq:mem_potentiali_LIF}, with observed neurons clamped to the training example $\bmx_{\leq t-1}$ and hidden neurons clamped to the samples ${\bmh}_{\leq t-1}$. Then, a central processor collects the log probabilities $p(x_{i,t} | {u}_{i,t})$ under the current model from all observed neurons $i \in \set{X}$ to compute the time-averaged learning signal ${\ell}_t$, as in \eqref{eq:eligibility-ls} and feeds back the global learning signal to all latent neurons. 

Intuitively, this learning signal indicates to the hidden neurons how effective their current signaling is in ensuring the desired input/output behavior with high probability. Finally, each observed and hidden neuron $i$ computes the eligibility trace $\bme_{i,t}$ of the gradient, i.e., $\grad_{\bmtheta_i} \log p_{\bmtheta_i}(x_{i,t} | \bmx_{\leq t-1}, {\bmh}_{\leq t-1})$ and $\grad_{\bmtheta_i} \log p_{\bmtheta_i}(h_{i,t} | \bmx_{\leq t-1}, {\bmh}_{\leq t-1})$, respectively, as in \eqref{eq:eligibility-grad-var}. The local parameters $\bmtheta_i$ of each observed neuron $i \in \set{X}$ are updated in the direction of the eligibility trace $\bme_{i,t}$, while each hidden neuron $i \in \set{H}$ updates the parameter using $\bme_{i,t}$ and the learning signal $\ell_t$ in \eqref{eq:eligibility-ls}.

{\bf Sparsity and regularization.} 
As discussed, the energy consumption of SNNs depends on the number of spikes emitted by the neurons. Since the ML criterion does not enforce any sparsity constraint, an SNN trained using the methods discussed so far may present dense spiking signals \cite{gerhard2017stability}. This is especially the case for the hidden neurons, whose behavior is not tied to the training data. To obviate this problem, it is possible to add a regularization term $-\alpha \cdot \text{{KL}} (q_\bmphi(\bmh_{\leq T} | \bmx_{\leq T})||r(\bmh_{\leq T}) )$ to the learning objective $L_{\bmx_{\leq T}}(\bmtheta,\bmphi)$ in \eqref{eq:elbo-general}, where $\text{KL}(p \parallel q) = \sum_{x} p(x) \log (p(x)/q(x))$ is the Kullback-Leibler divergence between distributions $p$ and $q$, $r(\bmh_{\leq T})$ represents a baseline distribution with the desired level of sparsity, and $\alpha > 0$ is a parameter adjusting the amount of regularization. This regularizing term, which penalizes variational distributions far from the baseline distribution, can also act as a regularizer to minimize overfitting by enforcing a bounded rationality constraint \cite{leibfried15:reward}. The learning rule in Algorithm~\ref{alg:latent-ml-vl} can be modified accordingly.

{\bf Interpretation.} 
The update \eqref{eq:latent-vl-sgd-online} for the synaptic weight $w_{j,i}$ of any observed neuron $i \in \set{X}$ follows the local two-factor rule, as described in the ``Interpretation'' section for fully observed models. In contrast, for any hidden neuron $i \in \set{H}$, the update applies a three-factor nonlocal learning rule \eqref{eq:three}. Accordingly, the postsynaptic error signal of hidden neuron $i$ and the filtered feedforward trace of presynaptic neuron $j$ are multiplied by the global learning signal \eqref{eq:ls-var}. As anticipated previously, the global learning signal can be interpreted as an internal reward signal. To see this more generally, we can rewrite \eqref{eq:ls-var} as 
\begin{align} \label{eq:ls-var-reward}
{\ell}_{\bmtheta^\textX}(\bmx_{\leq T}, {\bmh}_{\leq T}) = \log p_\bmtheta(\bmx_{\leq T} | {\bmh}_{\leq T}) - \log \frac{q_{\bmtheta^\textH}({\bmh}_{\leq T} | \bmx_{\leq T})}{p_\bmtheta({\bmh}_{\leq T})}.
\end{align}
According to \eqref{eq:ls-var-reward}, the learning signal rewards hidden spike signals ${\bmh}_{\leq T}$, producing observations $\bmx_{\leq T}$ that yield a large likelihood $\log p_\bmtheta(\bmx_{\leq T} | {\bmh}_{\leq T})$ for the desired behavior. Furthermore, it penalizes values of hidden spike signals ${\bmh}_{\leq T}$ that have large variational probability $q_{\bmtheta^\textH}({\bmh}_{\leq T} | \bmx_{\leq T})$ while having a low prior probability $p_\bmtheta({\bmh}_{\leq T})$ under the model. 

As discussed in the ``Learning Tasks'' section, SNNs can be trained in a batch or online mode. In the next sections, we provide a representative, simple, and reproducible example for each case.

\section*{Batch Learning Examples}


\begin{figure}[t!]
\centering
\includegraphics[height=0.25\columnwidth]{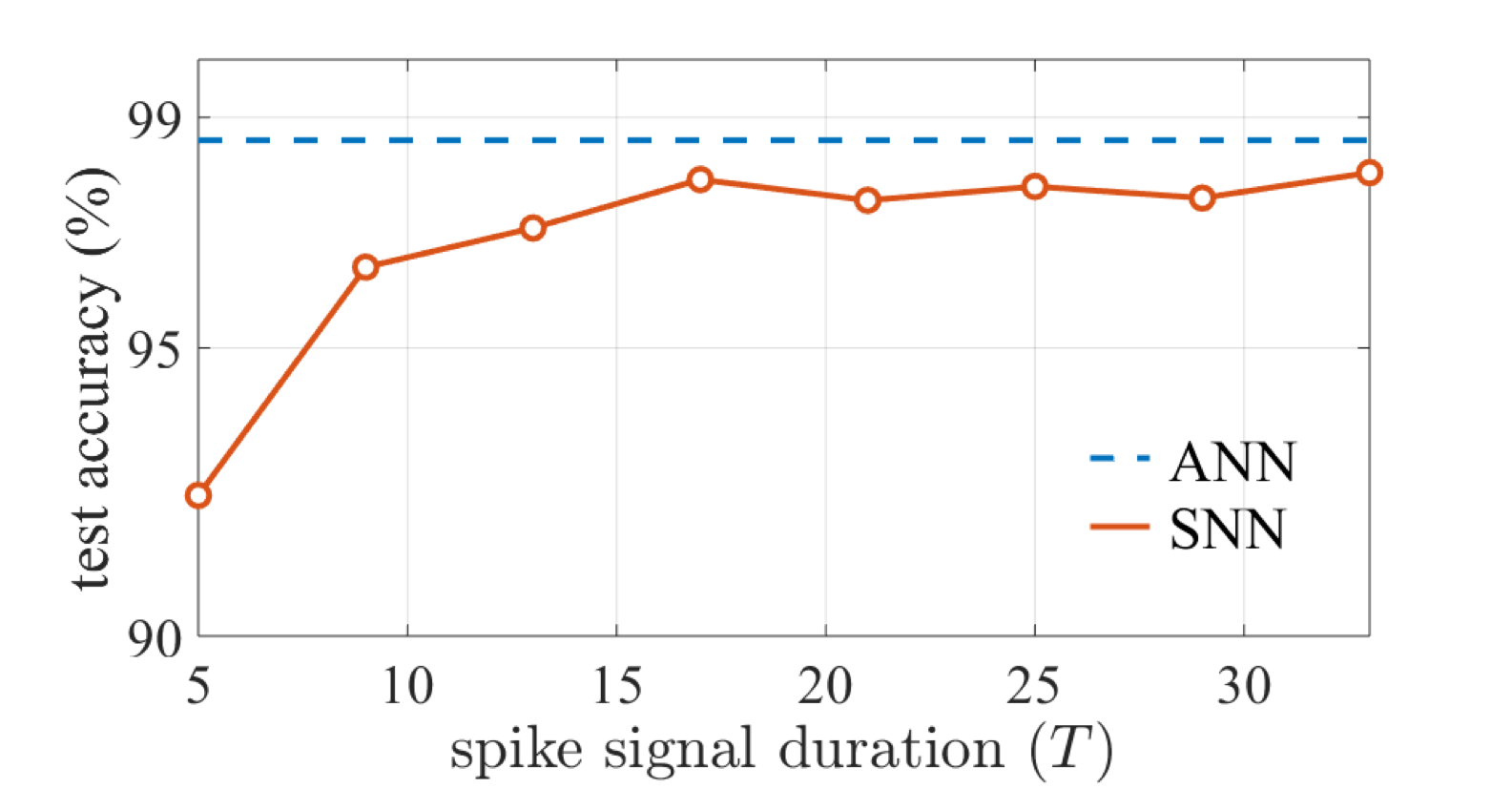}
\vspace{-0.2cm}
\caption{A performance of classification based on a two-layer SNN trained via batch ML learning in terms of accuracy versus the duration $T$ of the operation of the SNN. The performance of an ANN with the same topology is also shown as a baseline (see \cite{jang19:def} and \cite{jang19:spmsupp} for details).
}
\label{fig:ex_sup_fully_acc}
\vspace{-0.5cm}
\end{figure}

As an example of batch learning, we consider the standard handwritten digit classification task on the USPS data set \cite{hull1994usps}. We adopt an SNN with two layers, the first encoding the input and the second the output, with directed synaptic links existing from all neurons in the input layer to all neurons in the output layer. No hidden neurons exist, and, hence, training can be done as described in the section ``Training SNNs: Fully Observed Models''. Each $16 \times 16$ input image, representing either a ``$1$'' or a ``$7$'' handwritten digit, is encoded in the spike domain by using rate encoding. Each gray pixel is converted into an input spiking signal by generating an independent identically distributed (i.i.d.) Bernoulli vector of $T$ samples, with the spiking probability proportional to the pixel intensity and limited to between zero and $0.5$. As a result, we have $256$ input neurons, with one per pixel of the input image. The digit labels $\{1,7\}$ are also rate encoded using each one of the two output neurons. The neuron corresponding to the correct label index emits spikes with a frequency of one every three sample, while the other output neurons are silent. We refer the reader to \cite{jang19:def} and the supplementary material \cite{jang19:spmsupp} for further details on the numerical setup. 

Fig.~\ref{fig:ex_sup_fully_acc} shows the classification accuracy in the test set versus the duration $T$ of the operation of the SNN after the convergence of the training process. The classification accuracy of a conventional ANN with the same topology and a soft-max output layer is added for comparison. Note that, unlike the SNN, the ANN outputs real values, namely, the logits for each class processed by the soft-max layer. From the figure, the SNN is seen to provide a graceful tradeoff between accuracy and complexity of learning: as $T$ increases, the number of spikes that are processed and the output by the SNN grows larger, entailing a larger inference complexity but also an improved accuracy that tends to that of the baseline ANN.

\begin{figure}[t!]
\centering
\subfigure[]{
\includegraphics[height=0.3\columnwidth]{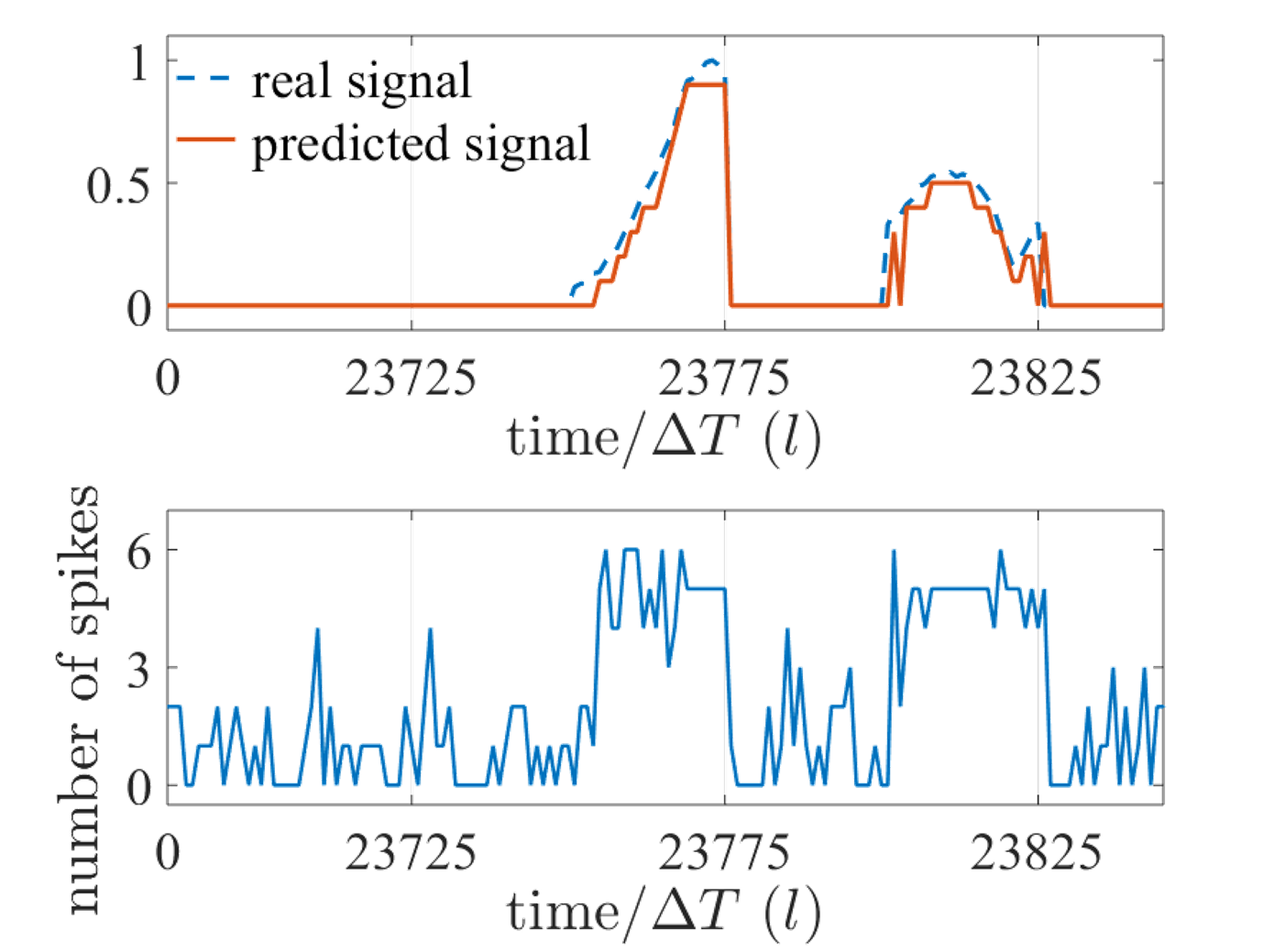} \label{fig:prediction_rate_raster_a}
} \hspace{-0.6cm}
\subfigure[]{
\includegraphics[height=0.3\columnwidth]{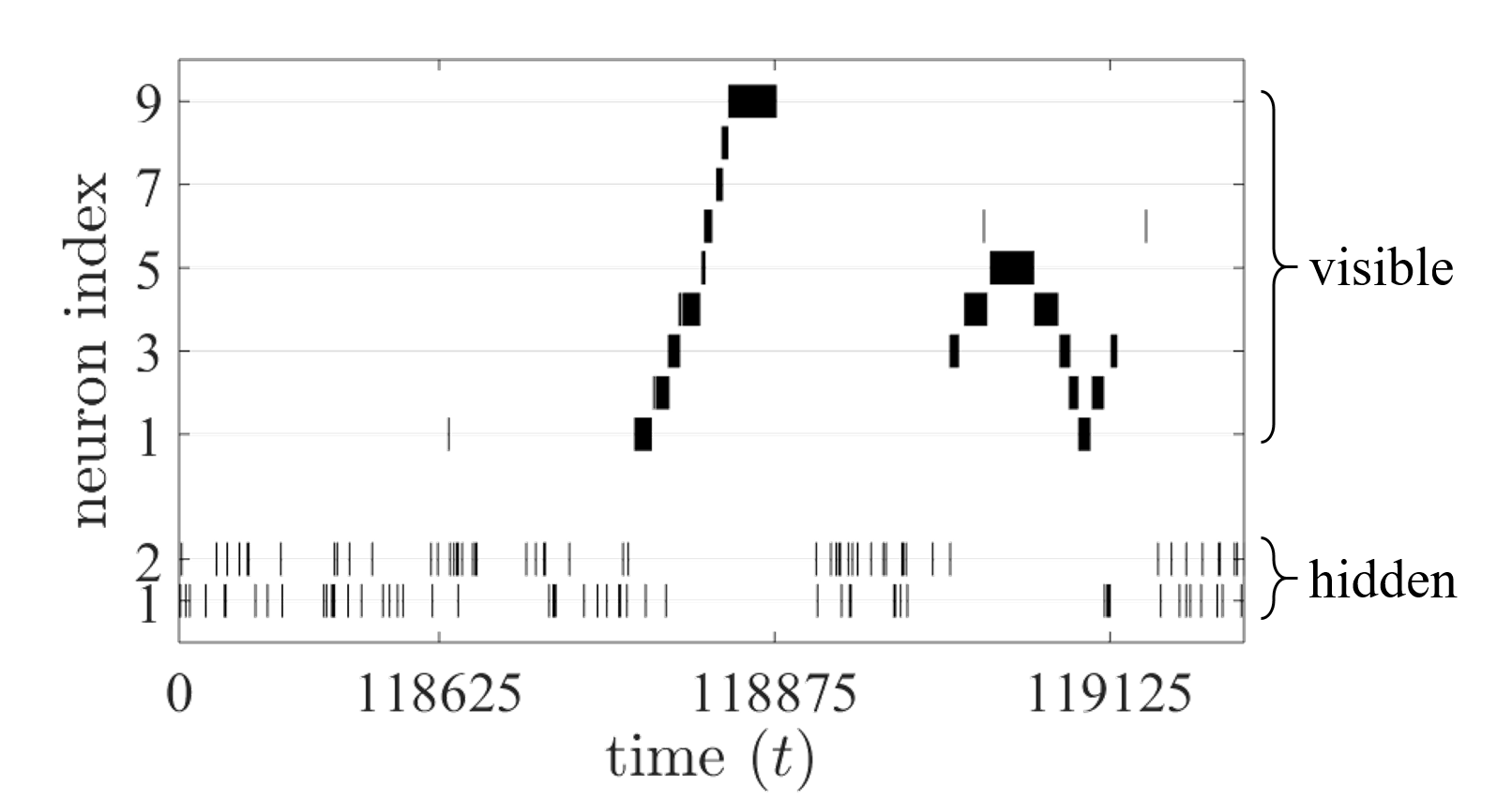} \label{fig:prediction_rate_raster_b}
}
  \caption{An online prediction task based on an SNN with $N_\textX = 9$ visible neurons and $N_\textH = 2$ hidden neurons trained via Algorithm~\ref{alg:latent-ml-vl}. (a) A real, analog time signal and a predicted, decoded signal (top) and the total number of spikes emitted by the SNN (bottom). (b) A spike raster plot of visible neurons (top) and a spike raster plot of hidden neurons (bottom). }
  \label{fig:prediction_rate_raster}
  \vspace{-0.55cm}
\end{figure}

\section*{Online Learning Examples}

We now consider an online prediction task in which the SNN sequentially observes a time sequence $\{a_l\}$ and the SNN is trained to predict, in an online manner, the next value of sequence $a_l$, given the observation of the previous values $a_{\leq l-1}$. The time sequence $\{a_l\}$ is encoded in the spike domain, producing a spike signal $\{\bmx_t\}$, consisting of $N_\textX$ spiking signals $\bmx_t = (x_{1,t},\ldots,x_{N_\textX,t})$ with $\Delta T \geq 1$ samples for each sample $a_l$. We refer to $\Delta T$ as a time expansion factor. Each of the spiking signals $x_{i,t}$ is associated with one of $N_\textX$ visible neurons. 

We adopt a fully connected SNN topology that also includes $N_\textH$ hidden neurons. In this online prediction task, we trained the SNN using Algorithm~\ref{alg:latent-ml-vl}, with the addition of a sparsity regularization term. This is obtained by assuming an i.i.d. reference Bernoulli distribution with a desired spiking rate $r \in [0,1]$, i.e., $ \log r(\bmh_{\leq T}) = \sum_{t=0}^T \sum_{i \in \set{H}} h_{i,t} \log r + (1-h_{i,t}) \log (1-r)$ (see the supplementary material \cite{jang19:spmsupp} for details). The source sequence is randomly generated as follows: at every $T_s = 25$ time steps, one of three possible sequences of duration $T_s$ is selected, namely, an all-zero sequence with probability $0.7$, a sequence of class $1$ from the SwedishLeaf data set of the UCR archive \cite{UCRArchive2018}, or a sequence of class $6$ from the same archive, with equal probability [see Fig.~\ref{fig:prediction_rate_raster_a} for an illustration].

\begin{figure}[t!]
\centering
\subfigure[]{
\includegraphics[width=0.45\columnwidth]{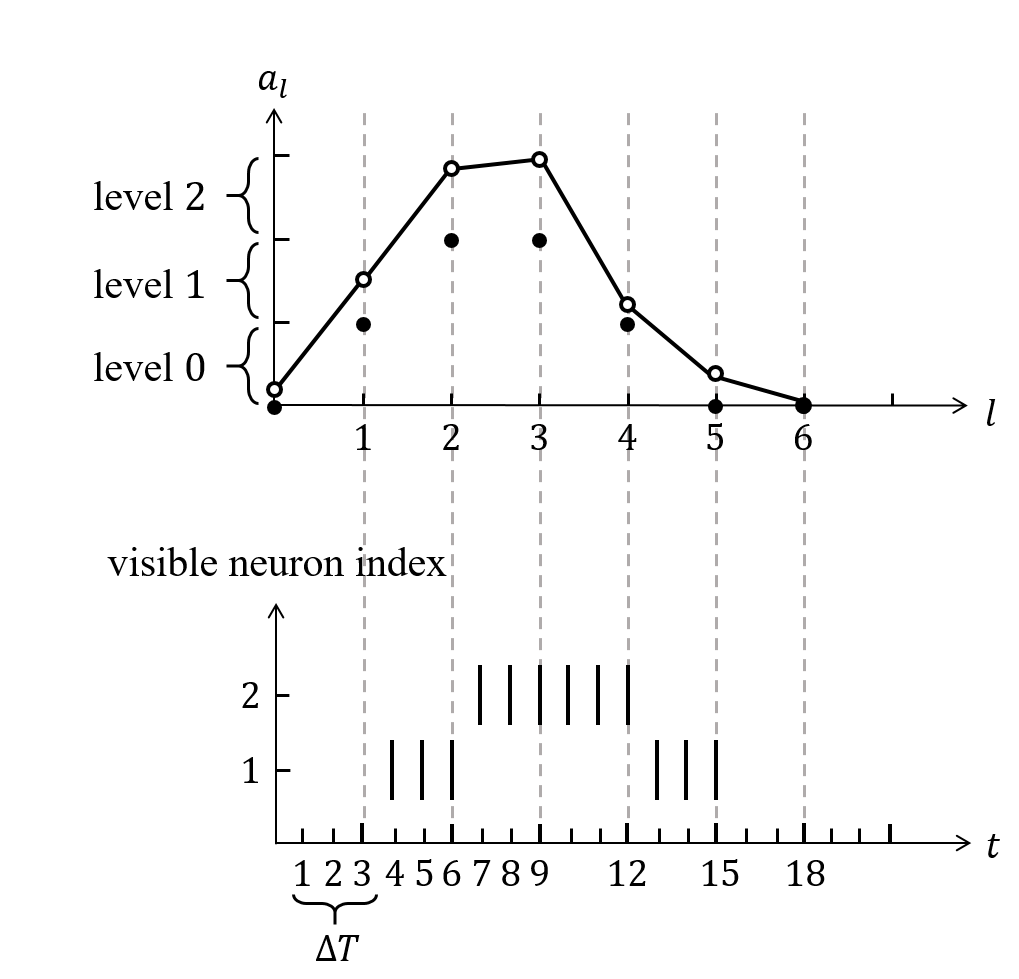} \label{fig:ex_coding_a}
} 
\subfigure[]{
\includegraphics[width=0.45\columnwidth]{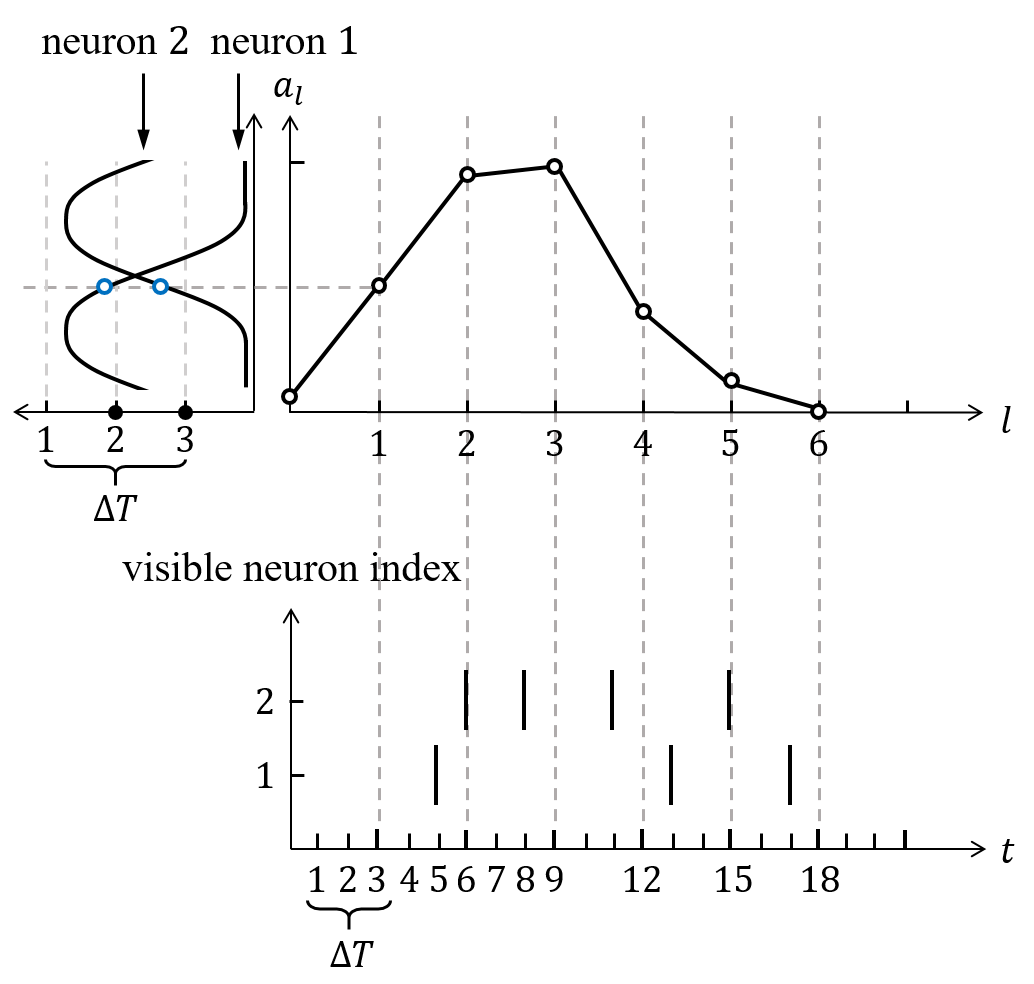} \label{fig:ex_coding_b}
} 
  \caption{Examples of coding schemes with $N_\textX = 2$ visible neurons and time expansion factor $\Delta T = 3$. (a) With rate coding, each value $a_l$ is discretized into $N_\textX + 1 = 3$ levels (top), and $\Delta T = 3$ consecutive spikes are assigned to input neuron $i$ for level $i = 1,2$, and no spikes are assigned otherwise (bottom). (b) With time coding, value $a_l$ is encoded for each visible neuron into zero or one spike, whose timing is given by the value of the corresponding Gaussian receptive field \cite{bohte2002unsupervised}. }
  \label{fig:ex_coding}
  \vspace{-0.55cm}
\end{figure}

{\bf Encoding and decoding.} 
Each value $a_l$ of the time sequence is converted into $\Delta T$ samples $\bmx_{l \Delta T + 1},$ $\bmx_{l \Delta T + 2}, \ldots, \bmx_{(l+1) \Delta T}$ of the $N_\textX$ spike signals $\{\bmx_t\}$ via rate or time coding, as illustrated in Fig.~\ref{fig:ex_coding}. With {\em rate coding}, the value $a_l$ is first discretized into $N_\textX + 1$ uniform quantization levels using rounding to the largest lower value. The lowest, ``silent'', level is converted to all-zero signals $\bmx_{l \Delta T + 1},\bmx_{l \Delta T + 2}, \ldots,$ $\bmx_{(l+1) \Delta T}$. Each of the other $N_\textX$ levels is assigned to a visible neuron, so that the neuron associated with the quantization level corresponding to value $a_l$ emits $\Delta T$ consecutive spikes while the other neurons are silent. Rate decoding predicts value $a_{l+1}$ by generating the samples $\bmx_{(l+1) \Delta T+1},\ldots,\bmx_{(l+2) \Delta T}$ from the trained model and then selecting the neuron with the largest number of spikes in this window. 

For {\em time coding}, each of the $N_\textX$ visible neurons is associated with a different shifted, truncated Gaussian receptive field \cite{bohte2002unsupervised}. Accordingly, as seen in Fig.~\ref{fig:ex_coding_b}, for each value $a_l$, each visible neuron $i$ emits a signal $x_{i, l \Delta T + 1}, x_{i, l \Delta T + 2}, \ldots, x_{i, (l+1) \Delta T}$ that contains no spike if the value $a_l$ is outside the receptive field and, otherwise, contains one spike, with the timing determined by the value of the corresponding truncated Gaussian receptive field quantized to values $\{1,\ldots, \Delta T\}$ using rounding to the nearest value. Time decoding considers the first spike timing of the samples $x_{i, (l+1) \Delta T + 1}, \ldots, x_{i, (l+2) \Delta T}$ for each visible neuron $i$ and predicts a value $a_{l+1}$ using a least-squares criterion on the values of the receptive fields (see \cite{eliasmith2004neural} and \cite{bohte2002unsupervised}). We refer to the supplementary material \cite{jang19:spmsupp} for further details on the numerical setup.

{\bf Rate coding.} 
First, assuming rate encoding with $\Delta T = 5$, we train an SNN with $N_\textX = 9$ visible neurons and $N_\textH = 2$ hidden neurons using Algorithm~\ref{alg:latent-ml-vl}. In the top portion of Fig.~\ref{fig:prediction_rate_raster_a}, we see a segment of the signal and of the prediction for a time window after the observation of the $23,700$ plus training samples of the sequence. The corresponding spikes emitted by the SNN [Fig.~\ref{fig:prediction_rate_raster_b}] are also shown, along with the total number of spikes per time instant [Fig.~\ref{fig:prediction_rate_raster_a}, bottom]. The SNN is seen to be able to provide an accurate prediction. Furthermore, the number of spikes, and, hence, the operating energy, depends on the level of activity of the input signal. This demonstrates the potential of SNNs for always-on event-driven applications. As a final note, in this particular example, the hidden neurons are observed to act as a detector of activity versus silence, which facilitates the correct behavior of the visible neurons. 

\begin{figure}[t!]
\centering
\includegraphics[height=0.26\columnwidth]{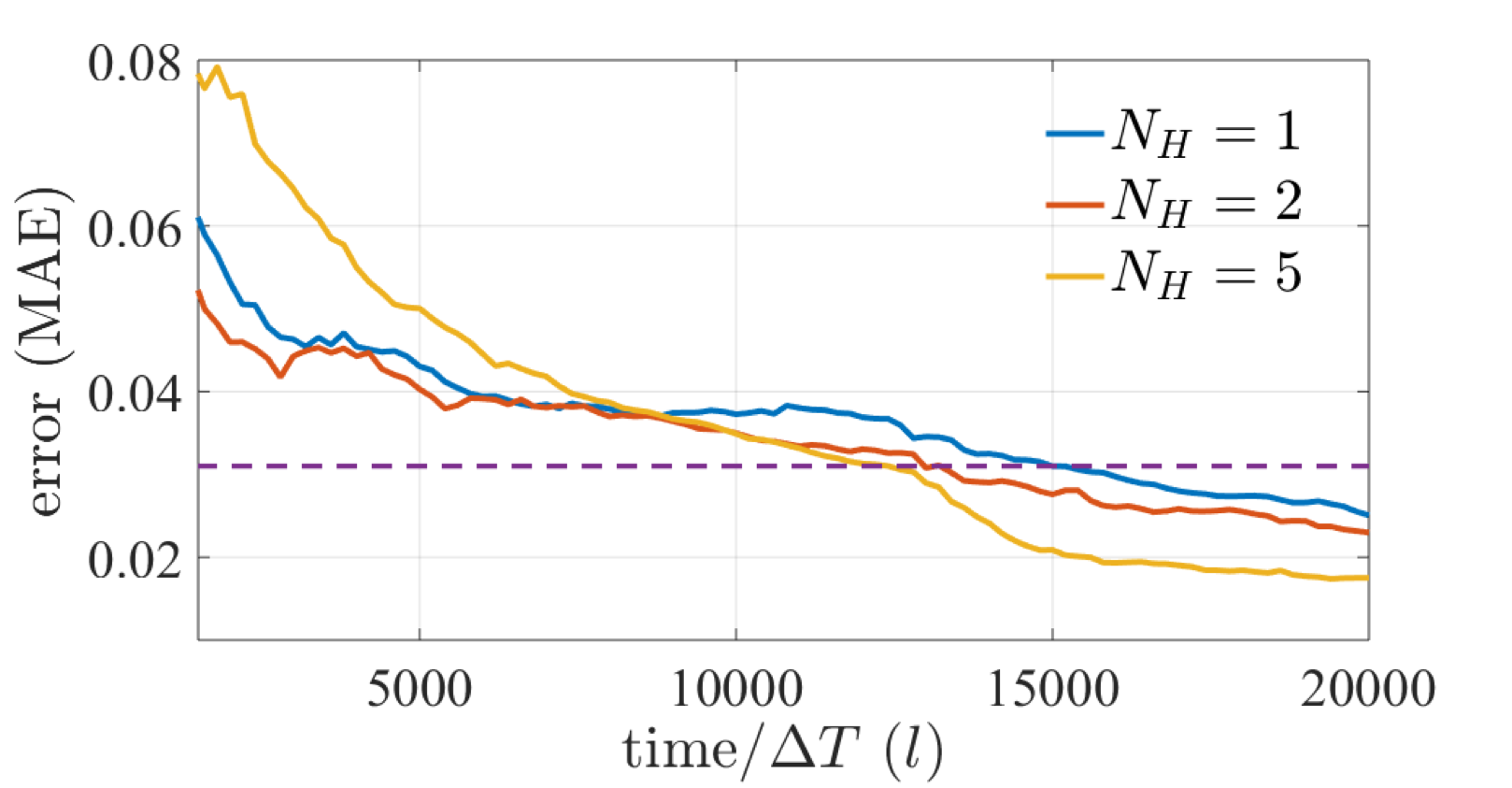}
\vspace{-0.2cm}
\caption{The prediction error versus training time for SNNs with $N_\textX = 9$ visible neurons and $N_\textH = 1,2,$ and $5$ hidden neurons trained via ML learning using Algorithm~\ref{alg:latent-ml-vl}. The dashed line indicates the performance of a baseline persistent predictor that outputs the previous sample (quantized to $N_\textX$ levels, as described in the text).
}
\label{fig:prediction_mae_hid}
\vspace{-0.5cm}
\end{figure}

The role of the number $N_\textH$ of hidden neurons is further investigated in Fig.~\ref{fig:prediction_mae_hid}, which shows the prediction error as a function of the number of observed training samples for different values of $N_\textH$. Increasing the number of hidden neurons is seen to improve the prediction accuracy as long as training is carried out for a sufficiently long time. The prediction error is measured in terms of average mean absolute error (MAE). For reference, we also compare the prediction performance with a persistent baseline (dashed line) that outputs the previous sample, upon quantization to $N_\textX$ levels for fairness.

{\bf Rate vs time encoding.} 
we now discuss the impact of the coding schemes on the online prediction task. We train an SNN with $N_\textX = 2$ visible neurons and $N_\textH = 5$ hidden neurons. Fig.~\ref{fig:prediction_rate_raster_a} shows the prediction error and Fig.~\ref{fig:prediction_rate_raster_b} the number of spikes in a window of $2,500$ samples of the input sequence, after the observation of the $17,500$ training samples, versus the time expansion factor $\Delta T$. From the figure, rate encoding is seen to be preferable for smaller values of $\Delta T$, while time encoding achieves better prediction error for larger $\Delta T$, with fewer spikes and, hence, energy consumption. 

This result is a consequence of the different use that the two schemes make of the time expansion $\Delta T$. With rate encoding, a larger $\Delta T$ entails a large number of spikes for the neuron encoding the correct quantization level, which provides increased robustness to noise. In contrast, with time encoding, the value $\Delta T$ controls the resolution of the mapping between input value $a_l$ and the spiking times of the visible neurons. This demonstrates the efficiency benefits of SNNs that may arise from their unique time encoding capabilities.

\begin{figure}[t!]
\centering
\subfigure[]{
\includegraphics[height=0.24\columnwidth]{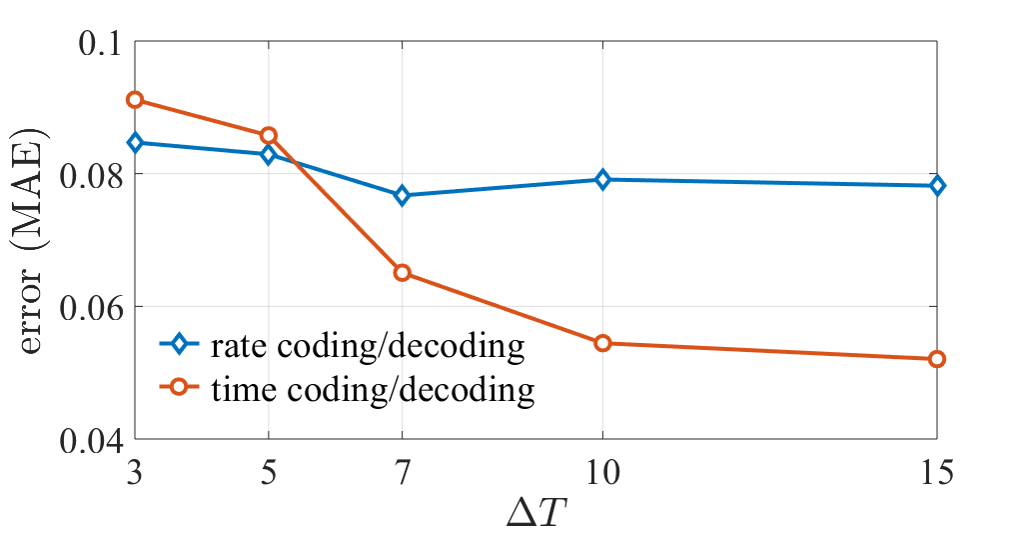}
} \hspace{-0.6cm}
\subfigure[]{
\includegraphics[height=0.24\columnwidth]{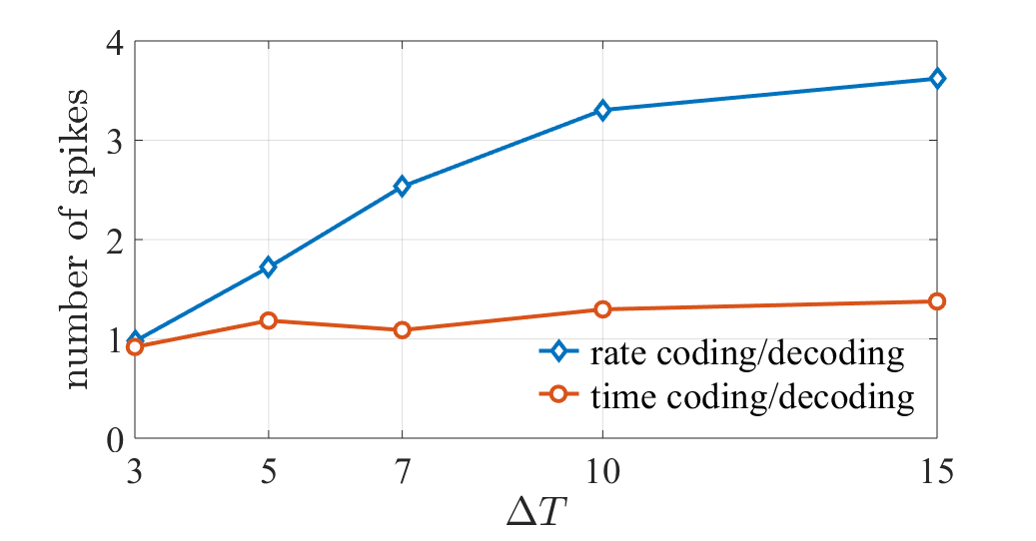}
}
  \caption{An online prediction task based on an SNN consisting of $N_\textX = 2$ visible neurons and $N_\textH = 5$ hidden neurons, with rate and time coding schemes: (a) the prediction error and (b) the number of spikes emitted by the SNN versus the time expansion factor $\Delta T$. }
  \label{fig:prediction_rate_time}
  \vspace{-0.5cm}
\end{figure}

\section*{Conclusions and Open Problems}\label{sec:conclusions}

As illustrated by the examples in the previous section, SNNs provide a promising alternative solution to conventional ANNs for the implementation of low-power learning and inference. When using rate encoding, they can approximate the performance of any ANN, while also providing a graceful tradeoff between accuracy, on the one hand, and energy consumption and delay, on the other. Most importantly, they have the unique capacity to process time-encoded information, yielding sparse, event-drive, and low-complexity inference and learning solutions.

The recent advances in hardware design reviewed in \cite{rajendran2019low} are motivating renewed efforts to tackle the current lack of well-established direct training algorithms that are able to harness the potential efficiency gains of SNNs. This article has argued that this gap is, at least in part, a consequence of the insistence on the use of deterministic models, which is in turn due to their dominance in the context of ANNs. As discussed, not only can probabilistic models allow the recovery of learning rules that are well known in theoretical neuroscience, but they can also provide a principled framework for the derivation of more general training algorithms. Notably, these algorithms differ significantly from the standard backpropagation approach used for ANNs, owing to their locality coupled with global feedback signaling. 

With the main aim of inspiring more research on the topic, this article has presented a review of models and training methods for probabilistic SNNs with a probabilistic signal processing framework. We focused on GLM spiking neuron models, given their flexibility and tractability, and on ML-based training methods. We conclude this article with some discussion on extensions in terms of models and algorithms as well as on open problem.

The SNN models and algorithms we have considered can be extended and modified along various directions. In terms of models, while randomness is defined here at the level of neurons' outputs, alternative models introduce randomness at the level of synapses or thresholds \cite{kasabov2010spike, mostafa2018learning}. Furthermore, while the models studied in this article encode information in the temporal behavior of the network within a given interval of time, information can also be retrieved from the asymptotic steady-state spiking rates, which define a joint probability distribution \cite{davies2018loihi, maass2014noise, bellec2018long}. Specifically, when the GLM \eqref{eq:joint-time}-\eqref{eq:prob-ind} has symmetric synaptic weights \note{-} i.e., $w_{j,i} = w_{i,j}$, the memory of the synaptic filter is $\tau = 1$, and there is no feedback filter, the conditional probabilities \eqref{eq:prob-ind} for all neurons define a Gibbs sampling procedure for a Boltzmann machine that can be used for this purpose. As another extension, more general connections among neurons can be defined, including instantaneous firing correlations, and more information, such as a sign, can be encoded in a spike \cite{jang19:def}. Finally, while here we focus on signal processing aspects, at a semantic level, SNNs can process logical information by following different principles \cite{eliasmith2004neural}.

In terms of algorithms, the doubly stochastic SGD approach reviewed here for ML training can be extended and improved by leveraging an alternative estimator of the ELBO and its gradients with respect to the variational parameters that is known as the {\em reparameterization trick} \cite{kingma14:vae}. Furthermore, similar techniques can be developed to tackle other training criteria, such as Bayesian optimal inference \cite{kappel15:synaptic}, reward maximization in reinforcement learning (see \cite{simeone2018brief} for a discussion in the context of general probabilistic models). 

Interesting open problems include the development of meta-learning algorithms, whereby the goal is learning how to train or adapt a network to a new task (see, e.g., \cite{bellec2018long}); the design of distributed learning techniques; and the definition of clear use cases and applications with the quantification of advantages in terms of power efficiency \cite{abr2018benchmarking}. Another important problem is the design of efficient input/ output interfaces between information sources and the SNN, at one end, and between the SNN and actuators or end users, on the other. In the absence of such efficient mechanisms, SNNs risk replacing the so-called memory wall of standard computing architectures with an input/output wall.

\section*{Acknowledgments}
This work was supported in part by the European Research Council under the European Union's Horizon 2020 research and innovation program under grant 725731 and by the U.S. National Science Foundation under grant ECCS 1710009. A.G (partly) and B.G (fully) are supported by the European Union's Horizon 2020 Framework Programme for Research and Innovation under the Specific Grant Agreement No. 785907 (Human Brain Project SGA2).



\bibliographystyle{IEEEtran}
\bibliography{refs}

\begin{thebibliography}{10}
\providecommand{\url}[1]{#1}
\csname url@samestyle\endcsname
\providecommand{\newblock}{\relax}
\providecommand{\bibinfo}[2]{#2}
\providecommand{\BIBentrySTDinterwordspacing}{\spaceskip=0pt\relax}
\providecommand{\BIBentryALTinterwordstretchfactor}{4}
\providecommand{\BIBentryALTinterwordspacing}{\spaceskip=\fontdimen2\font plus
\BIBentryALTinterwordstretchfactor\fontdimen3\font minus
  \fontdimen4\font\relax}
\providecommand{\BIBforeignlanguage}[2]{{%
\expandafter\ifx\csname l@#1\endcsname\relax
\typeout{** WARNING: IEEEtran.bst: No hyphenation pattern has been}%
\typeout{** loaded for the language `#1'. Using the pattern for}%
\typeout{** the default language instead.}%
\else
\language=\csname l@#1\endcsname
\fi
#2}}
\providecommand{\BIBdecl}{\relax}
\BIBdecl

\bibitem{welling18:icmltalk}
M.~Welling, ``Intelligence per kilowatt-hour,''
  \url{https://youtu.be/7QhkvG4MUbk}, 2018.

\bibitem{paugam2012computing}
H.~Paugam-Moisy and S.~Bohte, ``Computing with spiking neuron networks,'' in
  \emph{Handbook of Natural Computing}.\hskip 1em plus 0.5em minus 0.4em\relax
  Springer, 2012, pp. 335--376.

\bibitem{maass1997networks}
W.~Maass, ``Networks of spiking neurons: the third generation of neural network
  models,'' \emph{Neural networks}, vol.~10, no.~9, pp. 1659--1671, 1997.

\bibitem{davies2018loihi}
M.~Davies~et al., ``Loihi: A neuromorphic manycore processor with on-chip
  learning,'' \emph{IEEE Micro}, vol.~38, no.~1, pp. 82--99, 2018.

\bibitem{rajendran2019low}
B.~Rajendran, A.~Sebastian, M.~Schmuker, N.~Srinivasa, and E.~Eleftheriou,
  ``Low-power neuromorphic hardware for signal processing applications,''
  \emph{arXiv preprint arXiv:1901.03690}, 2019.

\bibitem{rueckauer2018conversion}
B.~Rueckauer and S.-C. Liu, ``Conversion of analog to spiking neural networks
  using sparse temporal coding,'' in \emph{Proc. IEEE International Symposium
  on Circuits and Systems (ISCAS)}, Florence, Italy, May 2018, pp. 1--5.

\bibitem{lee2016training}
J.~H. Lee, T.~Delbruck, and M.~Pfeiffer, ``Training deep spiking neural
  networks using backpropagation,'' \emph{Frontiers in Neuroscience}, vol.~10,
  p. 508, 2016.

\bibitem{OConnorW16}
P.~O'Connor and M.~Welling, ``Deep spiking networks,'' \emph{arXiv preprint
  arXiv:1602.08323}, 2016.

\bibitem{wu2018spatio}
Y.~Wu, L.~Deng, G.~Li, J.~Zhu, and L.~Shi, ``Spatio-temporal backpropagation
  for training high-performance spiking neural networks,'' \emph{Frontiers in
  Neuroscience}, vol.~12, 2018.

\bibitem{AbbotDayan}
P.~Dayan and L.~Abbott, \emph{Theoretical Neuroscience: Computational and
  Mathematical Modeling of Neural Systems}.\hskip 1em plus 0.5em minus
  0.4em\relax MIT Press, 2001.

\bibitem{eliasmith2004neural}
C.~Eliasmith and C.~H. Anderson, \emph{Neural engineering: Computation,
  representation, and dynamics in neurobiological systems}.\hskip 1em plus
  0.5em minus 0.4em\relax MIT press, 2004.

\bibitem{simeone2018brief}
O.~Simeone, ``A brief introduction to machine learning for engineers,''
  \emph{Foundations and Trends{\textregistered} in Signal Processing}, vol.~12,
  no. 3-4, pp. 200--431, 2018.

\bibitem{sutton2018reinforcement}
R.~S. Sutton and A.~G. Barto, \emph{Reinforcement learning: An
  introduction}.\hskip 1em plus 0.5em minus 0.4em\relax MIT press, 2018.

\bibitem{pillow08:spatio}
J.~W. Pillow~et al., ``Spatio-temporal correlations and visual signalling in a
  complete neuronal population,'' \emph{Nature}, vol. 454, no. 7207, p. 995,
  2008.

\bibitem{gerstner2002spiking}
W.~Gerstner and W.~M. Kistler, \emph{Spiking neuron models: Single neurons,
  populations, plasticity}.\hskip 1em plus 0.5em minus 0.4em\relax Cambridge
  University Press, 2002.

\bibitem{osogami17:BMtime}
T.~Osogami, ``Boltzmann machines for time-series,'' \emph{arXiv preprint
  arXiv:1708.06004}, 2017.

\bibitem{neal1992connectionist}
R.~M. Neal, ``Connectionist learning of belief networks,'' \emph{Artificial
  Intelligence}, vol.~56, no.~1, pp. 71--113, 1992.

\bibitem{gerhard2017stability}
F.~Gerhard, M.~Deger, and W.~Truccolo, ``On the stability and dynamics of
  stochastic spiking neuron models: Nonlinear hawkes process and point process
  glms,'' \emph{PLoS computational biology}, vol.~13, no.~2, p. e1005390, 2017.

\bibitem{gardner2015learning}
B.~Gardner, I.~Sporea, and A.~Gr{\"u}ning, ``Learning spatiotemporally encoded
  pattern transformations in structured spiking neural networks,'' \emph{Neural
  Computation}, vol.~27, no.~12, pp. 2548--2586, 2015.

\bibitem{neftci2019surrogate}
E.~O. Neftci, H.~Mostafa, and F.~Zenke, ``Surrogate gradient learning in
  spiking neural networks,'' \emph{arXiv preprint arXiv:1901.09948}, 2019.

\bibitem{fremaux2016neuromodulated}
N.~Fr{\'e}maux and W.~Gerstner, ``Neuromodulated spike-timing-dependent
  plasticity, and theory of three-factor learning rules,'' \emph{Frontiers in
  Neural Circuits}, vol.~9, p.~85, 2016.

\bibitem{brea2013matching}
J.~Brea, W.~Senn, and J.-P. Pfister, ``Matching recall and storage in sequence
  learning with spiking neural networks,'' \emph{Journal of Neuroscience},
  vol.~33, no.~23, pp. 9565--9575, 2013.

\bibitem{bleema2018fts}
B.~Rosenfeld, O.~Simeone, and B.~Rajendran, ``Learning first-to-spike policies
  for neuromorphic control using policy gradients,'' in \emph{Proc. IEEE
  International Workshop on Signal Processing Advances in Wireless
  Communications (SPAWC)}, Cannes, France, July 2019.

\bibitem{goodfellow2016deep}
I.~Goodfellow, Y.~Bengio, and A.~Courville, \emph{Deep learning}.\hskip 1em
  plus 0.5em minus 0.4em\relax MIT press, 2016.

\bibitem{bienenstock1982bcm}
E.~L. Bienenstock, L.~N. Cooper, and P.~W. Munro, ``Theory for the development
  of neuron selectivity: orientation specificity and binocular interaction in
  visual cortex,'' \emph{Journal of Neuroscience}, vol.~2, no.~1, pp. 32--48,
  1982.

\bibitem{watt10:homeostatic}
A.~J. Watt and N.~S. Desai, ``Homeostatic plasticity and {STDP}: keeping a
  neuron's cool in a fluctuating world,'' \emph{Frontiers in Synaptic
  Neuroscience}, vol.~2, p.~5, 2010.

\bibitem{mnih14:nvil}
A.~Mnih and K.~Gregor, ``Neural variational inference and learning in belief
  networks,'' in \emph{Proc. International Conference on Machine Learning
  (ICML)}, Beijing, China, June 2014, pp. 1791--1799.

\bibitem{kingma14:vae}
D.~P. Kingma and M.~Welling, ``Auto-encoding variational bayes,'' in
  \emph{Proc. International Conference on Learning Representations (ICLR)},
  Banff, Canada, Apr. 2014.

\bibitem{rezende14:vi_snn}
D.~J. Rezende and W.~Gerstner, ``Stochastic variational learning in recurrent
  spiking networks,'' \emph{Frontiers in Computational Neuroscience}, vol.~8,
  p.~38, 2014.

\bibitem{hinton00:sBM}
G.~E. Hinton and A.~D. Brown, ``Spiking boltzmann machines,'' in \emph{Proc.
  Advances in Neural Information Processing Systems (NIPS)}, Denver, US, Nov.
  2000, pp. 122--128.

\bibitem{kappel15:synaptic}
D.~Kappel, S.~Habenschuss, R.~Legenstein, and W.~Maass, ``Network plasticity as
  {B}ayesian inference,'' \emph{PLoS Computational Biology}, vol.~11, no.~11,
  p. e1004485, 2015.

\bibitem{leibfried15:reward}
F.~Leibfried and D.~A. Braun, ``A reward-maximizing spiking neuron as a bounded
  rational decision maker,'' \emph{Neural Computation}, vol.~27, no.~8, pp.
  1686--1720, 2015.

\bibitem{jang19:def}
H.~Jang and O.~Simeone, ``Training dynamic exponential family models with
  causal and lateral dependencies for generalized neuromorphic computing,'' in
  \emph{Proc. IEEE International Conference on Acoustics, Speech and Signal
  Processing (ICASSP)}, Brighton, UK, May 2019, pp. 3382--3386.

\bibitem{jang19:spmsupp}
H.~Jang, O.~Simeone, B.~Gardner, and A.~Gr{\"u}ning, ``An introduction to
  spiking neural networks: Probabilistic models, learning rules, and
  applications [supplementary material],''
  \url{https://nms.kcl.ac.uk/osvaldo.simeone/spm-supp.pdf}, 2019.

\bibitem{hull1994usps}
J.~J. Hull, ``A database for handwritten text recognition research,''
  \emph{IEEE Transactions on Pattern Analysis and Machine Intelligence},
  vol.~16, no.~5, pp. 550--554, 1994.

\bibitem{UCRArchive2018}
H.~A. Dau, E.~Keogh, K.~Kamgar, C.-C.~M. Yeh, Y.~Zhu, S.~Gharghabi, C.~A.
  Ratanamahatana, Yanping, B.~Hu, N.~Begum, A.~Bagnall, A.~Mueen, and
  G.~Batista, ``The ucr time series classification archive,''
  \url{https://www.cs.ucr.edu/~eamonn/time{\_}series{\_}data{\_}2018/}, 2018.

\bibitem{bohte2002unsupervised}
S.~M. Bohte, H.~La~Poutr{\'e}, and J.~N. Kok, ``Unsupervised clustering with
  spiking neurons by sparse temporal coding and multilayer rbf networks,''
  \emph{IEEE Transactions on Neural Networks}, vol.~13, no.~2, pp. 426--435,
  2002.

\bibitem{kasabov2010spike}
N.~Kasabov, ``To spike or not to spike: A probabilistic spiking neuron model,''
  \emph{Neural Networks}, vol.~23, no.~1, pp. 16--19, 2010.

\bibitem{mostafa2018learning}
H.~Mostafa and G.~Cauwenberghs, ``A learning framework for winner-take-all
  networks with stochastic synapses,'' \emph{Neural Computation}, vol.~30,
  no.~6, pp. 1542--1572, 2018.

\bibitem{maass2014noise}
W.~Maass, ``Noise as a resource for computation and learning in networks of
  spiking neurons,'' \emph{Proceedings of the IEEE}, vol. 102, no.~5, pp.
  860--880, 2014.

\bibitem{bellec2018long}
G.~Bellec, D.~Salaj, A.~Subramoney, R.~Legenstein, and W.~Maass, ``Long
  short-term memory and learning-to-learn in networks of spiking neurons,'' in
  \emph{Proc. Advances in Neural Information Processing Systems (NIPS)},
  Montreal, Canada, Dec. 2018, pp. 787--797.

\bibitem{abr2018benchmarking}
P.~Blouw, X.~Choo, E.~Hunsberger, and C.~Eliasmith, ``Benchmarking keyword
  spotting efficiency on neuromorphic hardware,'' \emph{arXiv preprint
  arXiv:1812.01739}, 2018.

\end{thebibliography}

\vskip -2\baselineskip plus -1fil

\begin{IEEEbiographynophoto}{Hyeryung Jang} (hyeryung.jang@kcl.ac.uk) received her B.S., M.S., and Ph.D. degrees in electrical engineering from the Korea Advanced Institute of Science and Technology, in 2010, 2012, and 2017, respectively. She is currently a research associate in the Department of Informatics, King's College London, United Kingdom. Her recent research interests lie in the mathematical modeling, learning, and inference of probabilistic graphical models, with a specific focus on spiking neural networks and communication systems. Her past research works also include network economics, game theory, and distributed algorithms in communication networks. 
\end{IEEEbiographynophoto}

\vskip -3\baselineskip plus -1fil

\begin{IEEEbiographynophoto}{Osvaldo Simeone} (osvaldo.simeone@kcl.ac.uk) received his M.Sc. degree (with honors) and Ph.D. degree in information engineering from Politecnico di Milano, Italy, in 2001 and 2005, respectively. He is a professor of information engineering with the Centre for Telecommunications Research, Department of Informatics, King's College London, United Kingdom. He is a corecipient of the 2019 IEEE Communication Society Best Tutorial Paper Award, the 2018 IEEE Signal Processing Society Best Paper Award, the 2017 Beest Paper by {\em Journal of Communications and Networks}, the 2015 IEEE Communication Society Best Tutorial Paper Award, and the IEEE International Workshop on Signal Processing Advances in Wireless Communications 2007 and IEEE Wireless Rural and Emergency Communications Conference 2007 Best Paper Awards. He currently serves on the editorial board of {\em IEEE Signal Processing Magazine} and is a Distinguished Lecturer of the IEEE Information Theorey Society. He is a Fellow of the Institution of Engineering and Technology and of the IEEE. 
\end{IEEEbiographynophoto}

\vskip -3\baselineskip plus -1fil

\begin{IEEEbiographynophoto}{Brian Gardner} (b.gardner@surrey.ac.uk) recieved his M.Phys. degree from the University of Exeter, United Kingdom, in 2011 and his Ph.D. degree in computational neuroscience from the University of Surrey, Guildford, United Kingdom, in 2016. He is a research fellow in the Department of Computer Science, University of Surrey. Currently, his research focuses on the theoretical aspects of learning in spiking neural networks. He is also working as a part of the Human Brain Project and is involved with the implementation of spike-based learning algorithms in neuromorphic systems for embedded applications. 
\end{IEEEbiographynophoto}

\vskip -3\baselineskip plus -1fil

\begin{IEEEbiographynophoto}{Andr\'e Gr\"uning} (andre.gruening@hochschule-stralsund.de) received his U.G. degree in Theoretical Physics from the University of G\''ottingen and his Ph.D. degree in Computer Science from the University of Leipzig. 
He is a professor of mathematics and computational intelligence at the University of Applied Sciences, Stralsund, Germany. He is a visiting member of the European Institute for Theoretical Neuroscience, Paris, France. Previously, he was a senior lecturer (associate professor) in the Department of Computer Science, University of Surrey, Guildford, United Kingdom. He held research posts in computational neuroscience at the Scuola Internazionale Superiore di Studi Avanzati, Trieste, Italy, and in cognitive neuroscience at the University of Warwick, Coventry, United Kingdom. His research concentrates on computational and cognitive neuroscience, especially learning algorithms for spiking neural networks. He is a partner in the Human Brain Project, a European Union Horizon 2020 Flagship Project. 
\end{IEEEbiographynophoto}

\end{document}